\documentclass[12pt,english]{article}
\usepackage[T1]{fontenc}
\usepackage[latin9]{inputenc}
\usepackage{geometry}
\usepackage{color}
\usepackage{array}
\usepackage{verbatim}
\usepackage{rotating}
\usepackage{multirow}
\usepackage{amsmath}
\usepackage{amsthm}
\usepackage{graphicx}
\usepackage{setspace}
\usepackage[authoryear]{natbib}
\makeatletter
\pdfoutput=1

\providecommand{\tabularnewline}{\\}

\theoremstyle{plain}
\newtheorem{thm}{\protect\theoremname}
\theoremstyle{plain}
\newtheorem{prop}[thm]{\protect\propositionname}
\theoremstyle{remark}
\newtheorem{rem}[thm]{\protect\remarkname}
\theoremstyle{plain}
\newtheorem*{lem*}{\protect\lemmaname}

\usepackage{graphicx,sectsty,amsthm,amssymb,multirow,amsmath}
\usepackage{epstopdf}
\usepackage{multirow}
\usepackage{bbm}
\usepackage{setspace}
\usepackage[ruled]{algorithm2e}
\date{}

\setcitestyle{round}

\@ifundefined{showcaptionsetup}{}{%
 \PassOptionsToPackage{caption=false}{subfig}}
\usepackage{subfig}
\makeatother

\usepackage{babel}
\providecommand{\lemmaname}{Lemma}
\providecommand{\propositionname}{Proposition}
\providecommand{\remarkname}{Remark}
\providecommand{\theoremname}{Theorem}

\begin{document}
\title{$\textrm{\ensuremath{\mathrm{AKM^{2}D}}}$ : An Adaptive Framework
for Online Sensing and Anomaly Quantification}
\author{Hao Yan$^1$, Kamran Paynabar$^2$, Jianjun Shi$^2$}
\date{   $^1$School of Computing, Informatics, and Decision Systems Engineering, Arizona State University, Tempe, AZ 85281\\
$^2$ School of Industrial and Systems Engineering, Georgia Institute of Technology, Atlanta, Georgia
}

\maketitle
\begin{abstract}
In point-based sensing systems such as coordinate measuring machines
(CMM) and laser ultrasonics where complete sensing is impractical
due to the high sensing time and cost, adaptive sensing through a
systematic exploration is vital for online inspection and anomaly
quantification. Most of the existing sequential sampling methodologies
focus on reducing the overall fitting error for the entire sampling
space. However, in many anomaly quantification applications, the main
goal is to estimate sparse anomalous regions in the pixel-level accurately.
In this paper, we develop a novel framework named Adaptive Kernelized
Maximum-Minimum Distance $(\textrm{\ensuremath{\mathrm{AKM^{2}D)}}}$
to speed up the inspection and anomaly detection process through an
intelligent sequential sampling scheme integrated with fast estimation
and detection. The proposed method balances the sampling efforts between
the space-filling sampling (exploration) and focused sampling near
the anomalous region (exploitation). The proposed methodology is validated
by conducting simulations and a case study of anomaly detection in
composite sheets using a guided wave test.
\end{abstract}

\section{Introduction \label{sec: Introduction}}

Systematic exploration of large areas for anomaly quantification is
of particular importance in various applications including quality
inspection, sensor network design, and structural health monitoring,
etc (e.g., in airplane maintenance \citep{Wu1996}, wafer manufacturing
\citep{jin2012sequential} and additive manufacturing \citep{Gibson2010}).
For example, in metrology and non-destructive evaluation (NDE), various
point-based sensing systems are used for quality inspection and anomaly
quantification. Examples include touch-probe coordinate measuring
machines (CMM) used for measuring the dimensional accuracy \citep{simpson1992mechanical},
and non-destructive methods such as guided wave-field tests (GWT)
\citep{mesnil2014frequency} and laser ultrasonics \citep{aussel1989precision},
utilized for defect quantification in composite sheets.

Most point-based sensing systems are only capable of measuring one
point at a time. The algorithm provides a binary map to show which
pixels are anomalies. Given this binary map, the number, location,
shape, or other features of anomalous regions can be easily derived
by simply using the morphological operations in image processing.
This binary map can be used for (1) determine the number of defective
areas (damages or imperfections); (2) locate each defective areas;
and (3) identify the shape of each defective area. This will be used
to determine if a part needs to be repaired or discarded; and to identify
potential root causes in the part fabrication process. However, to
achieve the foregoing goals, a point-based sensing method requires
measuring a large number of points sequentially, which results in
a time-consuming procedure not scalable to the online inspection of
large areas. For example, using a touch-probe CMM, it may take more
than $8$ hours to measure one typical batch of wafers that includes
$400$ wafers of $11$'' diameters \citep{jin2012sequential}. Also,
using guided wave test, the high-resolution inspection of a composite
laminate of size $0.4$ m$^{2}$ may take up to $4$ hours \citep{mesnil2016sparse}
(the experimental setup is shown in Figure 1). However, due to the
fact that anomalies are often clustered and sparse, most of the sensing
points are actually irrelevant and quickly locate the important regions
is important. Therefore, one can use a sequential and adaptive sampling
strategy to reduce the measurement time or energy consumption by reducing
the number of sampled points.

Existing adaptive sampling/sensing strategies in the literature can
be classified into three groups: the multi-resolution grid strategy,
sequential design of experiments (SDOE), and representative points
selection. 1) The multi-resolution grid sensing has been widely used
in practice. It begins with sensing over a coarse (low-resolution)
grid to estimate the underlying functional mean (e.g., the image background
in 2D measurements) and find the rough locations of the anomalies.
Then, sensing is continued over a finer (high-resolution) grids around
the identified anomalies to estimate the anomaly shape and size. The
performance of this method depends on the predefined size of both
course grids and fine grids, which should be specified based on the
size and shape of anomalies. Since such information may not be available
in advance, this method may result in either over-sampling or poor
anomaly quantification caused by under-sampling. 2) Many sampling
strategies have been developed in the area of the design of computer
experiments for modeling for spatial profiles. For example, some existing
research focuses on selecting design points on the unit hypercube.
Space-filling designs such as Latin hypercube design (LHD) \citep{mckay1979comparison}
and its many variations (e.g. \citealt{owen1994controlling,joseph2008orthogonal,ye1998orthogonal})
are widely used and have useful space-filling properties \citep{stein1987large}.
Other popular designs such as the Halton sequence \citep{halton1964algorithm}
and Sobol sequence \citep{soboldistribution}, mostly applied in Quasi
Morte Carlo to evaluate numerical integrals, have nice properties
such as uniformity and low discrepancy \citep{sobol1976uniformly}.
The main problem is that these methods are not sequential and cannot
direct the sampling direction toward the region of interest (anomaly
region). To address this issue, there has been a large amount of work
in the literature suggesting various sequential design of computer
experiments methods (SDOE) for improving the model fitting of spatial
profiles \citep{welch1992screening,bernardo1992integrated,jones1998efficient,ranjan2008sequential,jin2012sequential}.
\citet{loeppky2010batch} classified current SDOE methods into model-based
and distance-based (space-filling). Model-based methods include maximizing
the expected improvement criterion \citep{ranjan2008sequential,jones1998efficient},
minimizing the prediction error, minimizing the variance of the parameter
estimates, e.g., D-optimal design \citep{de1995d}, and optimizing
a composite index \citep{jin2012sequential}. Among the distance-based
models, sequential LHD design \citep{xiong2009optimizing,kyriakidis2005sequential}
and sequential maximin design \citep{stinstra2003constrained,loeppky2009choosing}
are popular. \citet{loeppky2009choosing} and \citet{loeppky2010batch}
showed the latter methods perform well not only in selecting the initial
sampling points but also in determining the follow-up runs for sequential
design of experiments since one can place an upper bound on the MSE
based on the distance measure of the design. The main problem of SDOE
methods is that they only focus on improving the estimation of the
functional mean over the entire sampling space without considering
potential anomalies and non-smooth features. 3) \citet{joseph2015sequential}
proposed the minimum energy design that selects representative points
based on a known distribution over the design space and sequentially
chooses the next design points based on a criterion minimizing the
total potential energy. However, the main problem of applying this
approach for online anomaly quantification is that the anomaly distribution
is often unknown a priori. Therefore, it lacks the ability of focused
sampling near anomalous regions.

The second relevant body of literature deals with function estimation
in the presence of anomalies. Robust kernel regression \citep{zhu2008robust}
and robust spline estimation \citep{de1972calculating} are among
these methods. However, their main focus is the estimation of the
functional mean, not the anomaly, and hence, they do not consider
the spatial structure of anomalies. To address this issue, \citet{yan2015anomaly}
and \citet{yan2018real} proposed smooth-sparse decomposition (SSD)
for anomaly detection in spatial profiles and temporal profiles, respectively.
SSD can separate anomalies from the functional mean by utilizing the
spatial structure of both the functional mean and anomalies. SSD,
however, can only work when measurements are dense, hence, not applicable
in point-based sensing and inspection systems.

The third relevant body of literature deals with the adaptive sampling
problem for change point detection or statistical process control
(SPC). For example, \citet{li2014statistical} proposed a dynamic
sampling scheme for SPC based on the p-value of the conventional CUSUM
control chart. \citet{liu2015adaptive} and \citet{wang2018spatial}
proposed an adaptive sampling method to dynamically update the sampling
location based on the current CUSUM statistics. However, the major
focus of SPC is to detect the out-of-control samples as soon as possible.
The accurate quantification of the anomaly is not the primary focus
for the SPC applications. Even though many SPC methods \citep{liu2015adaptive,zou2009multivariate,wang2018spatial}
have the capability of identifying the location of the anomaly, it
is typically not possible to fully quantify the anomaly when it is
first detected. In comparison, the proposed method focuses on the
online anomaly quantification with the least number of samples to
achieve simultaneous anomaly quantification when it is first detected.
In this case, we first need to detect the anomaly location and then
to use more samples to provide accurate quantification. Online anomaly
quantification is typically performed after the SPC to provide accurate
anomaly quantification. Furthermore, SPC focuses on timely detection
of a change in a dynamic setting. We assume the sample background
and anomaly does not change over time.

The main objective of this paper is to propose a new adaptive sampling
framework along with estimation procedures for online anomaly quantification.
The immediate benefit of the proposed framework is to help scale up
point-based sensing methods so that they can be used for in-situ inspection.
It can also be used in the sensing point selection for online anomaly
detection in the sensor network \citep{wang2004entropy}. An effective
adaptive sensing strategy should consist of two major elements: first,
it should randomly search the entire space (exploration) to spot anomalous
regions and recover the functional mean; and second, it should perform
the focused sampling of the areas near the anomalous regions (exploitation)
to determine the size and the shape of anomalies. To achieve this,
the following two challenges should be addressed: 1) how to intelligently
decide on the location of the next sampled point; and 2) how to estimate
anomalous regions as well as the functional mean online based on the
sparsely sampled points. In this paper, we will address the first
challenge by proposing a new sensing strategy named Adaptive Kernelized
Maximum Minimum-Distance $(\textrm{\ensuremath{\mathrm{AKM^{2}D)}}}$
combining the computer design of experiment approach for the random
exploration of the entire space and the Hilbert Kernel approach \citep{devroye1999hilbert}
for the focused sampling in anomalous regions (exploitation). We also
show the relationship of the proposed method to the existing sequential
design of experiment methods. To address the second challenge about
anomaly estimation, we propose a modeling framework based on robust
kernel regression for estimating the background (profile mean) and
sparse kernel regression for estimating and separating anomalies.
In order to perform both estimation and adaptive measurement in real-time,
we also propose efficient optimization algorithms.

\begin{onehalfspace}
The remainder of the paper is organized as follows. Section 2 provides
an overview of the proposed methodology. In Section 3, we propose
the new adaptive sampling/sensing framework $\textrm{\ensuremath{\mathrm{AKM^{2}D}}}$.
Section 4 elaborates mean estimation and anomaly detection algorithms.
In Sections 5 and 6, simulated data and a case study of anomaly detection
in composite laminates are used to evaluate the performance of the
proposed methodology. Finally, we conclude the paper with a short
discussion and directions for future work in Section 7.
\end{onehalfspace}

\section{Methodology Overview}

We first briefly review the overall methodology proposed in this paper.
We assume that the quality measure can be modeled by a true background
function $\mu(r)$ and the anomaly function $a(r)$, and the measurement
at location $r$ is given as 
\[
y(r)=\mu(r)+a(r)+\epsilon(r).
\]
The goal is to estimate the background function $\mu(r)$ and anomaly
function $a(r)$. This is different from SDOE methods, where the main
goal is to estimate the mean function. Here we also assume that the
function $\mu(r)$ and $a(r)$ are static and do not change over time.
Therefore, to achieve this goal, our methodology should include an
adaptive sampling framework to identify the best location of the next
point as well as an estimation procedure for recovering the background
function and quantifying anomalies. The adaptive sampling framework
is to decide the sampling location $r_{n+1}$ based on observations
$y(r)$ at $r_{1},\cdots,r_{n}$. The estimation procedure should
be able to update functional estimation on $\mu(r)$ and $a(r)$ given
a new observation $r_{n+1}$. In this paper, we assume that the function
background $\mu(r)$ is smooth and the anomaly $a(r)$ is sparse and
clustered. Finally, we assume that the measurement noises $\epsilon(r)$
in different locations are independent and follow the normal distribution
$N(0,\sigma^{2})$.

For illustration purposes and simplicity, we use the $2D$ sampling
space $\lbrack0,1\rbrack^{2}$ in this paper and further constrain
the samples to be on a $2D$ fine grid defined as $\mathcal{G}_{m}=\{(\frac{i}{m},\frac{j}{m})|i,j=1,\cdots,m\}$,
where $m$ can be specified by the resolution capability of the sensing
device. In the application of spatial sensing in composite part inspection,
typically the dimension does not exceed $2$ (e.g., $x,y$ axis).
However, the proposed method can be extended to a high-dimensional
sampling space. More discussion is added in the methodology section.

\begin{figure}
\centering\includegraphics[width=1\linewidth]{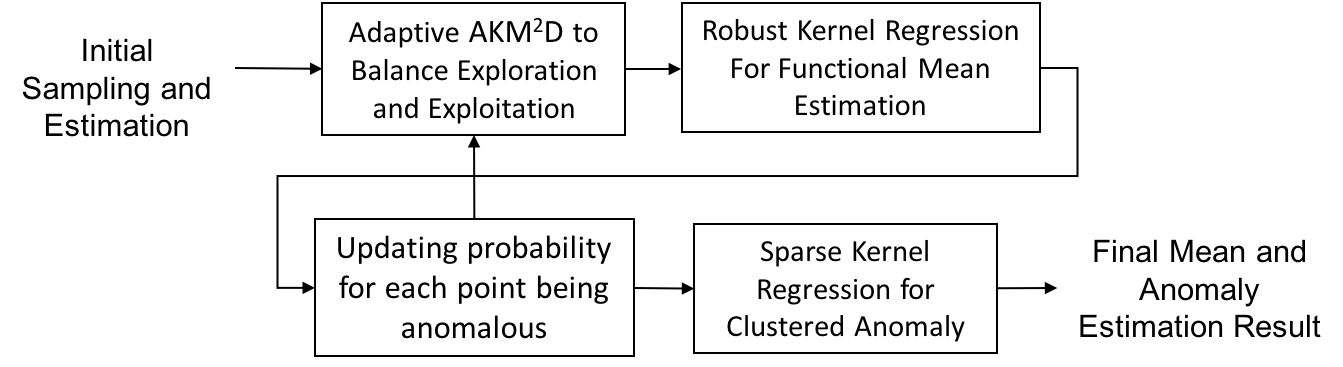}

\caption{Procedure of the proposed sampling algorithm}

\label{Fig: Procedure}
\end{figure}

The proposed methodology, illustrated in Figure \ref{Fig: Procedure},
is summarized as follows: First, $n_{init}$ initial points are sampled
using a space-filling design (e.g. max-min distance, \citep{johnson1990minimax})
to explore the entire sampling space. Then, based on the outcome of
the initial points, subsequent points are chosen by using $\textrm{\ensuremath{\mathrm{AKM^{2}D}}}$
to balance between the space-filling sampling (exploration) and the
focused sampling near the anomalous region (exploitation). After $\textrm{\ensuremath{\mathrm{AKM^{2}D}}}$
chooses the location of a new sample, the functional mean is estimated
(updated) via the robust kernel regression. After a certain number
of sampled points, if the functional mean estimate does not deviate
much from the estimate obtained in the previous iteration, the functional
mean estimation step can be skipped to reduce the computational time.
The probability that a point in the sample space is anomalous is then
updated after the estimation step of the functional mean, which is
used as an input for $\textrm{\ensuremath{\mathrm{AKM^{2}D}}}$ in
the next iteration. Next, clustered anomalous regions are estimated
(updated) via the proposed sparse kernel regression. Finally, this
procedure is repeated until the desired sampling resolution is reached.
The sampling resolution can be defined as the maximin distance, which
can be defined as the maximum distance of points in the entire sampling
space to the nearest sampled point.

In developing the proposed sampling methodology, we make the following
assumptions: we assume that sparse anomalies are in the form of clusters.
Also, for estimating the functional mean and anomalous regions, it
is assumed that the functional mean is smooth and anomalies have different
intensity values from the functional mean. It should be noted that
the proposed $\textrm{\ensuremath{\mathrm{AKM^{2}D}}}$ framework
is general and does not require the smoothness assumption.

\section{Adaptive Kernelized Maximum Minimum-Distance $\mathbf{(\textrm{\ensuremath{\mathrm{AKM^{2}D)}}}}$
Sensing Algorithm}

\subsection{Formulation and Algorithm}

In this section, we present our new adaptive sensing framework, $\textrm{\ensuremath{\mathrm{AKM^{2}D}}}$,
that helps sequentially choose the location of samples. Denote $\mathcal{M}_{n}$
as the observed samples in the $n^{th}$ iterations, defined as $\{r_{k}=(x_{k},y_{k})\in\mathcal{G}_{m}|k=1,\cdots,n\}$
are observed. Let $p_{a}(r_{k})$ denote the known probability that
the point $r_{k}$ in this set is anomalous (the detailed procedure
for estimating $p_{a}(r_{k})$ will be discussed in Section 4.) To
find the next sampled point $r_{n+1}$, we propose the following criterion:
\begin{equation}
r_{n+1}=\arg\max_{r}g_{n}(r)=\arg\max_{r}\psi_{n}(r)(f_{n}(r))^{\lambda},\label{eq: fmed}
\end{equation}
where $\psi_{n}(r)$ is the estimated distribution of anomalies. Therefore,
maximizing $\psi_{n}(r)$ can encourage the focused sampling (exploitation)
meaning that the next sampled point $r_{n+1}$ continues searching
in anomalous regions. $f_{n}(r)$ can be understood as the regularization
term to prevent sampled points being too close to each other. In other
words, $f_{n}(r)$ encourage the exploration of the entire sampling
space for undiscovered anomalies (space-filling property). In this
paper, we define $\psi_{n}(r)$ as a mixture distribution of Gaussian
distributions centered at each anomalous point observed and a uniform
distribution for the entire sampling space to account for unobserved
anomalies. That is, $\psi_{n}(r)=(\sum_{k=1}^{n}p_{a}(r_{k})K_{h}(r,r_{k})+u)$
where $K_{h}(r,r_{k})=\frac{1}{(\sqrt{2\pi}h)^{2}}\exp(-\frac{\|r-r_{k}\|^{2}}{2h^{2}})$
is the 2D-Gaussian kernel centered at point $r_{k}$ used to model
the clustered structure of the anomalies. $p_{a}(r_{k})$ and $u$
are respectively the mixture weights for the Gaussian distribution
$K_{h}(r,r_{k})$ and the uniform distribution. Note that $p_{a}(r_{k})$
is also the probability that the sampled point $r_{k}$ is anomalous.
Even though the normalization weight $\frac{1}{\sum_{k=1}^{n}p_{a}(r_{k})+u}$
changes over different iterations, it can still be neglected since
it is independent of $r$ and doesn't affect the optimization result
in \eqref{eq: fmed}. Furthermore, we define $f_{n}(r)$ by $f_{n}(r):=\min_{r_{k}\in\mathcal{M}_{n}}\|r-r_{k}\|$
to encourage the space-filling property. For a special case $\psi_{n}(r)=1$,
Equation \eqref{eq: fmed} becomes $r_{n+1}=\arg\max_{r}\min_{k=1,\cdots,n}\|r-r_{k}\|$
which is equivalent to a greedy approach to solve the maximum minimum-distance
design proposed by \citet{johnson1990minimax}. By plugging in $\psi_{n}(r)$
and $f_{n}(r)$ , the sampling criterion given in \eqref{eq: fmed}
can be rewritten as 
\begin{equation}
r_{n+1}=\mathrm{argmax}_{r}\left\{ (\sum_{k=1}^{n}p_{a}(r_{k})K_{h}(r,r_{k})+u)\min_{k=1,\cdots,n}\|r-r_{k}\|^{\lambda}\right\} .\label{eq: med}
\end{equation}
To efficiently solve \eqref{eq: med} on $r\in\mathcal{G}_{m}$, we
compute $\psi_{n}(r)$ by the tensor product of two 1D-Gaussian kernels.
That is, $\Psi_{n}=K_{x}^{T}P_{A}K_{y}+u1_{m\times m}$, where $K_{x,ij}=K_{y,ij}=\frac{1}{(\sqrt{2\pi}h)^{2}}\exp(-\frac{\|i-j\|^{2}}{2h^{2}m^{2}})$,
and $P_{A,ij}=p_{a}(\frac{i}{m},\frac{j}{m})1((\frac{i}{m},\frac{j}{m})\in\mathcal{M}_{n})$
are the $(i,j)$ component of the matrix $K_{x},K_{y}$ and $P_{A}$,
respectively. $1(x)$ is an indicator function defined as $1(x)=\begin{cases}
1 & x\text{ is true}\\
0 & x\text{ is false}
\end{cases}$, and $1_{m\times m}$ is an $m$ by $m$ matrix of $1$s. It is straightforward
to show that $f_{n}(r),r\in\mathcal{G}_{m}$ can be updated recursively
by $f_{n}(r)=\min(f_{n-1}(r),\|r-r_{n}\|),r\in\mathcal{G}_{m}$. Both
the space and time complexity of this recursive update is $O(m^{2})$,
where $m$ is the grid size in each dimension. Therefore, \eqref{eq: med}
can be efficiently and recursively solved by Algorithm \ref{alg: medgrid}.
Here, we would like to emphasize that the current optimization algorithm
is based on a grid search approach, which the function value of the
grid $\mathcal{G}_{m}$ is updated in each iteration. To optimize,
the largest numeric value of the function $f(r)$ on the grid $\mathcal{G}_{m}$
can be computed. However, we found that this grid searching algorithm
is actually more efficient than the global solver in the 2D sampling
space as it is a recursive algorithm and hence it requires updating
the function value for a small portion of the points. However, this
approach is only feasible if the dimension of the sampling space is
not too large. To optimize the design in a higher dimensional sampling
space, global optimization solver such as the particle swarm algorithm
or the genetic algorithm can be used. For more details about using
these solvers to find the near-optimal design, please refer to the
following paper \citep{mak2018minimax}.

\begin{algorithm}[tbp] 
\caption{AKMMD} 
\SetKwBlock{Initialize}{initialize}{end}
\Initialize{

Initial $n_{init}$ sampling based on max-min distance design

}

\For{$n=n_{init},\cdots,n_{max}$}{

Update $\psi_{n}(r)$ based on $\Psi_{n}=K_{x}^{T}P_{A}K_{y}+u1_{m\times m}$

Update $f_{n}(r)=\min(f_{n-1}(r),\|r-r_{n}\|)$ for $r\in\mathcal{G}_{m}$

$r_{n+1}=\mathrm{argmax}_{r\in\mathcal{G}_{m}}\psi_{n}(r)(f_{n}(r))^{\lambda}$

}

\label{alg: medgrid}

\end{algorithm}

\subsection{$\mathbf{\textrm{\ensuremath{\mathrm{AKM^{2}D}}}}$ Sampling Properties}

\begin{figure}
\centering\includegraphics[width=0.8\linewidth]{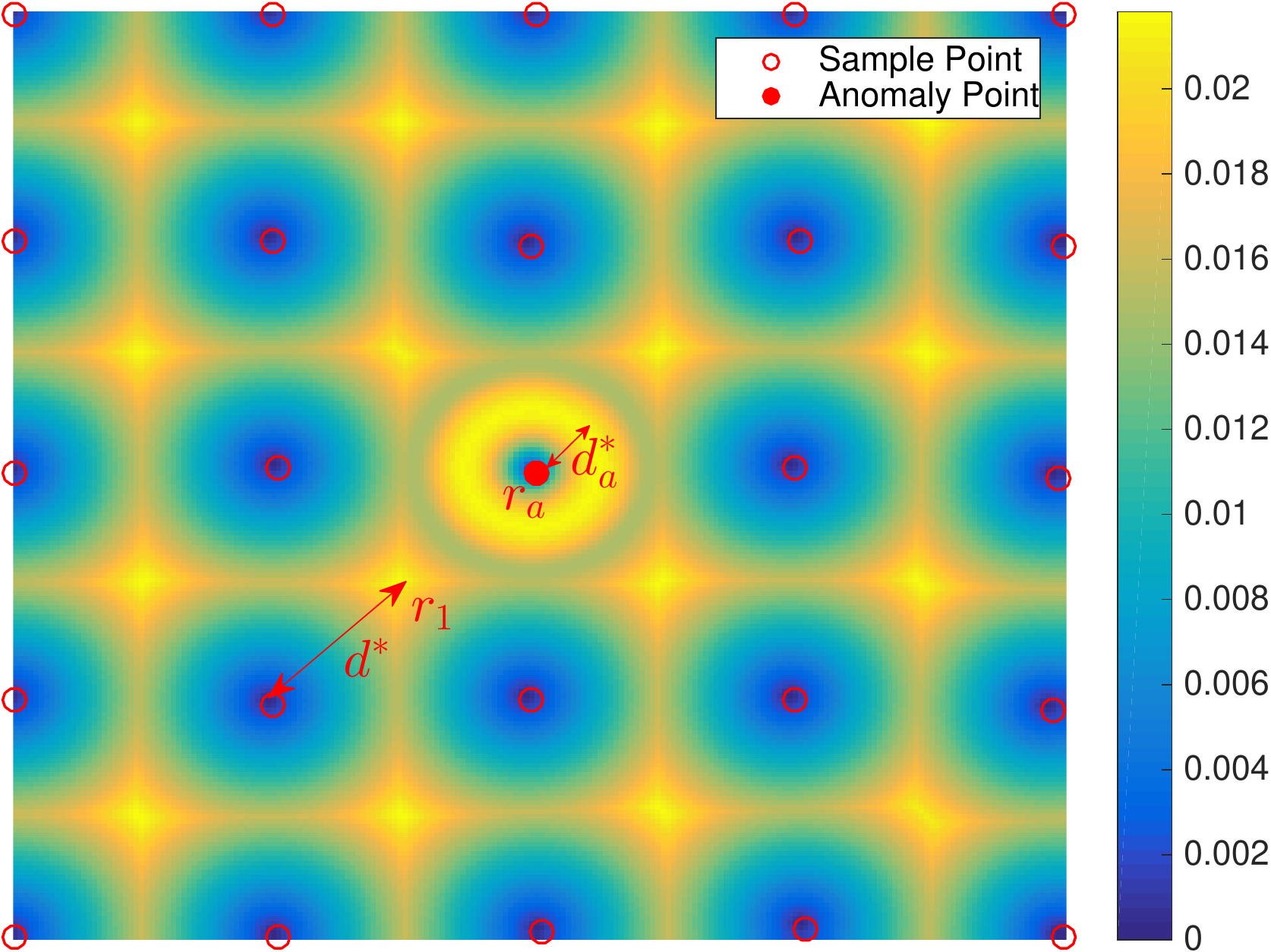}

\caption{The behavior of $g(r)$ with the center point as the anomaly point}

\label{Fig: miny}
\end{figure}

In this section, we study the properties of the proposed $\mathrm{\mathbf{\textrm{\ensuremath{\mathrm{AKM^{2}D}}}}}$.
Let $\mathcal{R}_{i}$ denote the neighborhood of a point $r_{i}$
defined by $\mathcal{R}_{i}=\{r|\|r-r_{i}\|\leq\|r-r_{k}\|,\forall k=1,\cdots,n\}$.
We first investigate the behavior of the sampling criterion $g(r)$
in the neighborhood of an anomalous point $r_{a}$, i.e., $\mathcal{R}_{a}$.
(see Figure \ref{Fig: miny}). It is easy to show that Equation \eqref{eq: med}
for $r\in\mathcal{R}_{a}$ can be decomposed into two terms: $g(r)=g_{a}(r)+g_{-a}(r)$,
where $g_{a}(r)=(K_{h}(r,r_{a})+u)\|r-r_{a}\|^{\lambda}$, $g_{-a}(r)=\left(\sum_{k\not=a}^{n}p(r_{k})K_{h}(r,r_{k})\right)\|r-r_{a}\|^{\lambda}$.
The second term, $g_{-a}(r)$, is often negligible in the neighborhood
of $r_{a}$ especially when $\|r_{k}-r_{a}\|\gg h,\forall k\neq a$.
For simplicity, in this subsection, we assume $r_{a}$ is the only
detected anomalous point with $p_{a}>0$.
\begin{prop}
The local maximum of $g_{a}(r),r\in\mathcal{R}_{a}$ is attained at
$\|r-r_{a}\|=d_{a}^{*}=h\sqrt{\lambda-2W(-\frac{\pi h^{2}\lambda u}{p_{a}}\exp(\frac{\lambda}{2}))}$
if $\{r:\|r-r_{a}\|=d_{a}^{*}\}\in\mathcal{R}_{a}$. $W$ is the Lambert
W-function defined as $W(z)=\{w|z=w\exp(w)\}$. \label{prop: medradius}
\end{prop}

The proof is given in Appendix A.

Proposition \ref{prop: medradius} guarantees that $g_{a}(r)$ in
the neighborhood of $r_{a}$ will generate a local maximum ring with
radius $d_{a}^{*}$ (as shown in Figure \ref{Fig: miny}), which encourages
the next sampled point to be chosen near the potential anomalous point
$r_{a}$ (exploitation), but with the distance of $d_{a}^{*}$ to
avoid over-exploitation. Proposition \ref{prop: medradius} only guarantees
the local optimality. However, the next sampled point is selected
on the local maximum ring only if it is the global maximum of $g(r)$.
To study this and show how criterion \eqref{eq: med} is able to balance
sampled points between exploration and exploitation, we give the following
necessary condition under which the algorithm selects $r_{a}^{*}$
.
\begin{prop}
Let $d^{*}$ denote the current sampling Max-Min Distance (MMD) defined
as the maximum distance of each point in the entire sampling space
with its closest sampled point, i.e., $d^{*}:=\max_{r}\min_{r_{k}\in\mathcal{M}_{n}}\|r-r_{k}\|$
and suppose $r_{a}$ is the only sampled point with $p_{a}>0$. $\|r-r_{a}\|=d_{a}^{*}=h\sqrt{\lambda-2W(-\frac{\pi h^{2}\lambda u}{p_{a}}\exp(\frac{\lambda}{2}))}$
is the global maximum of \eqref{eq: med} if \label{prop:  MEDbalance}
\begin{equation}
d^{*}<\tilde{d}^{*}:=(\frac{1}{(1+\exp(-c^{2})}\times\frac{(d_{a}^{*})^{2}}{2((d_{a}^{*})^{2}-\lambda h^{2})})^{\frac{1}{\lambda}}d_{a}^{*},\label{eq: ratio}
\end{equation}
where $c$ is a constant satisfies $c<\min_{r_{k}\neq r_{a},r_{k}\in\mathcal{M}_{n}}\|r_{k}-r_{a}\|/2\sqrt{2h^{2}\ln(\frac{p_{a}}{2\pi h^{2}u})})$.
\end{prop}

The proof is given in Appendix B.

Proposition \ref{prop:  MEDbalance} shows that the proposed algorithm
first samples the entire sampling space up to a certain resolution
$\tilde{d}^{*}$ and then, starts focused sampling. This ensures that
the proposed method does not miss any anomaly with the radius greater
than $\tilde{d}^{*}$. Furthermore, this proposition can be used for
choosing the tuning parameters, which will be discussed in the next
section.

To illustrate the implication of this proposition, we plot the behavior
of $g(r)$ in Figure \ref{Fig: miny}. The center point in this figure
is an anomalous point (the point indicated by $r_{a}$), which generates
a local optimal ring with a radius $d_{a}^{*}$. It will be global
optimum if this optimal value is larger than the other local maximum
in the center of the potential sampled points (the point indicated
by $r_{1}$) as shown in Figure \ref{Fig: miny}. Proposition \ref{prop:  MEDbalance}
shows that if \eqref{eq: ratio} holds, the algorithm will select
a point on the local maximum ring centered at $r_{a}$ as the global
optimum and hence as the next sampled point.

Proposition 1 and Proposition 2 study the transient behavior of the
algorithm and how it balances the exploration and exploitation when
the anomaly is first discovered. We also study the limiting behavior
of the proposed sampling algorithm in Proposition 3 and Remark 4.
\begin{prop}
\textup{If $\psi_{n}(r)=1$, the sampling points distribution will
converge to the uniform distribution. Furthermore, the Max-Min Distance
(MMD) of the sampling points decrease at the rate of $O(\frac{1}{n^{1/p}})$
for a $p-$dimensional sampling space.}
\end{prop}

The proof of Proposition 3 follows directly by Theorem 1 in the Minimum
Energy Design \citep{joseph2015sequential}. Proposition 3 demonstrates
the limiting distribution of the proposed algorithm when the anomaly
is not presented (i.e. $\psi_{n}(r)$ is a constant) is actually the
uniform distribution. Furthermore, even though this rate $O(\frac{1}{n^{1/p}})$
can only be proved when $\psi_{n}(r)=1)$, it is proved in \citep{joseph2015sequential},
$O(\frac{1}{n^{1/p}})$ is actually the upper bound of the convergence
rate for any design. We found in our simulation study that the MMD
also decrease at the rate of $O(\frac{1}{n^{1/p}})$. This result
can guide the practitioners on deciding the number of points needed
for the sampling algorithm.
\begin{rem}
If the $\psi_{n}(r)$ converges to $\psi(r)$, with proper initial
sampling points, the limiting distribution of the sampling algorithm
$r_{n+1}=\arg\max_{r}\{\psi(r)\min\|r-r_{k}\|^{\lambda}\}$ will converge
to $\psi(r)^{p/\lambda}$, where $p$ is the dimension of the sampling
space.
\end{rem}

Remark 4 is a conjecture and depends on whether the limiting behavior
of Minimum Energy Design \citep{joseph2015sequential} can be proved.
The assumptions of Remark 4 are discussed in Appendix C.

Remark 4 demonstrates the limiting behavior of the proposed algorithm.
It demonstrates that the limiting distribution is related to the function
$\psi_{n}(r)$. Recall that $\psi_{n}(r)=(\sum_{k=1}^{n}p_{a}(r_{k})K_{h}(r,r_{k})+u)$,
which is a combination of the background and anomaly. Therefore, the
proposed algorithm is able to balance exploration and exploitation.
Finally, as the number of sampling points $n\rightarrow\infty$, the
kernel density estimation $\frac{1}{n}\sum_{k=1}^{n}p_{a}(r_{k})K_{h}(r,r_{k})$
will approach the true anomaly distribution. In the limiting behavior,
when $n\rightarrow\infty$, more samples will be put on the anomaly
regions due to the increasing weight according to the $\sum_{k=1}^{n}p_{a}(r_{k})K_{h}(r,r_{k})$
due to the sample size.

\subsection{Tuning Parameter Selection}

In this section, we will discuss how these propositions can help us
select the tuning parameters $\lambda,h$ and $u$. First, based on
our numerical experiments in the simulation study, we suggest the
kernel bandwidth $h$ is selected approximately at the scale of the
desired anomaly sampling resolution (e.g., AMMD).

In the simulation, we find out $\lambda$ needs to be at least $5$
for the algorithm to demonstrate the behavior of both exploration
and exploitation. correspondingly, $u$ is normally set small as $u<10^{-7}.$
To decide the exact value of $\lambda$ and $u$, we can use the transient
behavior of the algorithm demonstrated in Propositions 1 and 2 to
select the tuning parameter. Since $d_{a}^{*}=h\sqrt{\lambda-2W(-\frac{\pi h^{2}\lambda u}{p_{a}}\exp(\frac{\lambda}{2}))}\approx h\sqrt{\lambda}$
when $u$ is small, $h\sqrt{\lambda}$ should be roughly the desired
AMMD for exploiting the anomaly. Furthermore, according to Proposition
2, since $d_{a}^{*}-\lambda h^{2}\approx2\frac{\pi h^{2}\lambda u}{p_{a}}\exp(\frac{\lambda}{2}))$,
we know $d^{*}=(4\pi u\exp(\frac{\lambda}{2}))^{-1/\lambda}h\sqrt{\lambda}$.
Note that when computing $\tilde{d}^{*}$, we ignore $(\frac{1}{1+\exp(-c^{2})})^{\frac{1}{\lambda}}$
since it is close to $1$ when $c>3$ and $\lambda>5$. For example,
$c=3$, $\lambda=5$, $(\frac{1}{1+\exp(-c^{2})})^{\frac{1}{\lambda}}=0.99998$.
If $d^{*}$ is larger than the size of sampling space (i.e. $d^{*}>1$),
the algorithm may be trapped in the anomalous region since it may
never start exploration. Therefore, we can set $(4\pi u\exp(\frac{\lambda}{2}))^{-1/\lambda}h\sqrt{\lambda}<1$.
for example, if the desired $h=0.02$ and we set $\lambda=5$, $u=1e-9$,
this inequality implies $\lambda$ should be at least $5$.

\subsection{Relationship with model-based criterion in SDOE}

In this section, we aim to link the proposed criterion in \ref{eq: fmed}
with the existing sequential design of experiment (SDOE) methods and
show how it can be derived from the SDOE perspective. We can use a
Gaussian process on the entire sampling space to represent the mean
response $\mu\sim GP(0,K(\cdot))$, with $K$ matrix defined as $K_{ij}=K(r_{i}-r_{j})=\exp(-\frac{\|r_{i}-r_{j}\|^{2}}{h^{2}})$.
The anomaly can be defined as the location where the measurement $y$
deviate from the mean response $\mu$ (i.e. $y-\mu$ is large). Equivalently,
the anomaly should be located in locations where the mean squared
prediction error $MSE(r)$ is large. It is straightforward to prove
that for a Gaussian process $MSE(r)\leq c_{0}\min_{k=1,\cdots,n}\|r-r_{k}\|^{2}$
\citep{loeppky2010batch}, where $c_{0}$ is a constant. Since we
only care about the anomalous regions, we define $IMSE=\left[\int\{\sqrt{MSE(r,D,\theta)}\psi_{n}(r)\}^{\beta}dr\right]^{1/\beta}$,
as the integrated MSE over the anomalous density, where $\psi_{n}(r)$
is the estimated distribution of anomaly $r$. IMSE can be also explained
as the integrated confidence intervals of the prediction.

Note that when $\beta\rightarrow\infty$, $IMSE\leq\max_{r}\min_{k=1,\cdots,n}\|r-r_{k}\|\psi_{n}(r)$,
and therefore, the solution of $r_{n+1}=\arg\max_{r}\min_{k=1,\cdots,n}\|r-r_{k}\|\psi_{n}(r)$
gives the point which contributes most to reducing the integrated
confidence intervals of the prediction. This provides an SDOE-based
justification for the proposed sampling criterion and strategy in
\eqref{eq: fmed}. The only difference between IMSE-based and $\mathrm{\mathbf{\textrm{\ensuremath{\mathrm{AKM^{2}D}}}}}$
criteria is the parameter $\lambda$ that is used to adjust the relative
importance of the $\psi_{n}(r)$ compared to the exploration part.

\section{Mean and Anomaly Estimation Using Sparse Samples\label{sec: Methodology}}

In the previous section, we proposed a general adaptive sampling strategy
and discussed its properties. Here, we propose methods for estimating
the mean function as well as anomalous regions using the sparse measurements
obtained by $\mathrm{\mathbf{\mathrm{AKM^{2}D}}}$. Specifically,
we present a robust kernel regression algorithm for functional mean
estimation and a sparse kernel regression algorithm for anomaly estimation.

\subsection{Robust Kernel Regression for Functional Mean Estimation}

Let $z_{k}$ denote the recorded measurement at point $r_{k}=(x_{k},y_{k})$
and $z=(z_{1},\cdots,z_{k},\cdots,z_{n})$ be the vector of measurements
for all $n$ sampled points. To model the smooth functional mean $\mu$
in the presence of anomalies, Reproducing Kernel Hilbert Space (RKHS)
is utilized. From the representer theorem \citep{scholkopf2001generalized},
it is known that every function in an RKHS can be written as a linear
combination of kernel functions evaluated at sampled points. If anomalies
did not exist, kernel regression could be used for estimating the
functional mean. However, since anomalies have a different functional
structure from the mean, they behave as outliers when estimating the
functional mean. Therefore, we utilize robust kernel regression to
alleviate the effect of anomalies on mean estimation. To estimate
the functional mean $\mu$, we minimize 
\begin{equation}
\sum_{k=1}^{n}\rho(z_{k}-\mu_{k})+\lambda\|\mu\|_{H},\label{eq: RKR}
\end{equation}
in which $\rho(x)$ is the Huber loss function, defined by $\rho(x)=\begin{cases}
x^{2} & |x|\leq\frac{\gamma}{2}\\
\gamma|x|-\frac{\gamma^{2}}{4} & |x|>\frac{\gamma}{2}
\end{cases}$, and $\lambda\|\mu\|_{H}$ is the Hilbert norm penalty, which controls
the smoothness of the functional mean. The Robust kernel regression
can be solved efficiently via an iterative soft-thresholding function
\citep{Mateos2012}. See Appendix C for the detailed derivation and
optimization algorithm. The functional mean $\mu$ is almost the same
after sensing enough sampled points. Therefore, to speed up the algorithm,
we stop updating $\mu$ when the estimation difference after adding
a new sampled point is smaller than a certain threshold. After estimating
the functional mean $\mu_{k}$, the residuals can be computed by $\hat{e}=[\hat{e}_{k}]=[z_{k}-\hat{\mu}_{k}]$.

\subsection{Updating Probability $p_{a}(r_{k})$}

We conduct a hypothesis test on the residual $\hat{e}_{k}$ to test
whether there exist anomalies in the specimen at the location $r_{k}$.
The null hypothesis is $H_{0}:\mu_{e_{k}}=0$, implying no anomalies
exist. The p-value of this test can be used to update the probability
of the sampled point $r_{k}$ being anomalous. That is, $p_{a}(r_{k})=P(|e_{k}|>|\hat{e}_{k}||e_{k}\sim N(0,\hat{s}^{2}))=1-2P(e_{k}>\hat{e}_{k})=2\Phi(\frac{|\hat{e}_{k}|}{\hat{s}})-1$,
where $\Phi(\cdot)$ is the cumulative density function of the standard
normal distribution, $\hat{s}$ is the standard deviation of the noise
$e$, which can be estimated by the median absolute deviation under
the normality assumption as $\hat{s}=\mathrm{median}\{|\hat{e}|\}/0.6745$.
$p_{a}(r_{k})$ is used as an input to $\textrm{\ensuremath{\mathrm{AKM^{2}D}}}$
as discussed earlier. Moreover, the selection of $\gamma$ can be
determined based on a specified false positive rate, $\alpha_{0}$,
associated with the hypothesis test. If no anomalies exist ($H_{0}$
is true), the false positive rate can be computed by $P(|e_{k}|>\frac{\gamma}{2}|e_{k}\sim N(0,\hat{s}^{2}))=2(1-\Phi(\frac{\gamma}{2\hat{s}}))=\alpha_{0}$.
Consequently, $\gamma$ can be selected by $\hat{\gamma}=2\hat{s}\Phi^{-1}(1-\frac{\alpha_{0}}{2})$.
See Appendix C for the reason as to why $\frac{\gamma}{2}$ is a good
threshold to determine whether a point is anomalous.

\subsection{Sparse Kernel Regression for Clustered Anomaly Estimation}

In this subsection, we estimate the size, shape, and boundary of anomalous
regions. Specifically, we model the spatial structure of clustered
anomalies by a Gaussian kernel $K_{a}$ through optimizing
\begin{equation}
\arg\min_{\theta_{a}}\|\hat{e}-K_{a}\theta_{a}\|^{2}+\gamma_{a}|\theta_{a}|_{1}.\label{eq: sparsekernel}
\end{equation}
Problem \eqref{eq: sparsekernel} can be solved efficiently by existing
L1 solvers such as the accelerated proximal gradient (APG). The APG
algorithm for solving Problem \eqref{eq: sparsekernel} is given in
Algorithm \ref{alg: sparsekernel}. For the tuning parameter $\gamma_{a}$,
as it has been pointed out by \citet{yan2015anomaly}, Generalized
Cross Validation (GCV) usually tends to select more points, leading
to a larger false positive rate. Therefore, instead of using GCV,
we choose $\gamma_{a}$ based on a specified false positive rate $\alpha$.
Since there is no closed-form solution for Problem \eqref{eq: sparsekernel}
with general $K_{a}$, Monte Carlo simulations can be used to select
$\gamma_{a}$ as follows: We first generate white noise from $e\sim NID(0,\hat{s}^{2})$,
where $\hat{s}$ is the standard deviation of the noise $e$. We then
select $\gamma_{a}$ such that $\alpha\times100\%$ of $\hat{a}=K_{a}\hat{\theta}_{a}$
are non-zero. Note that since $K_{a}$ changes overtime, $\gamma_{a}$
should be recomputed whenever a new point is measured, which is time-consuming.
Therefore, an approximate procedure for tuning parameter selection
is proposed. When $K_{a}$ is orthogonal, $\theta_{a}$ has a closed-form
solution computed by $\hat{\theta}_{a}=S_{\frac{\gamma_{a}}{2}}(K_{a}^{T}\hat{e})$,
or equivalently, $\hat{\theta}_{ai}=S_{\frac{\gamma_{\alpha}}{2}}(\sum_{j}K_{a}(r_{j},r_{i})\hat{e}_{j})$.
When $K_{a}$ is close to orthogonal, the soft-thresholding function
gives a reasonable approximate solution. The false positive rate can
then be computed by $\alpha=P(\hat{\theta}_{ai}\neq0)=2P(|z|>\frac{\gamma_{a}}{2}|z\sim N(0,l^{2}\hat{s}^{2})=2\Phi(1-\frac{\gamma_{a}}{2l\hat{s}})$,
where $l^{2}=\sum_{j}K_{a}(r_{j},r_{i})^{2}$. Therefore, $\gamma_{a}$
can be approximated by $\gamma_{a}=2l\hat{s}\Phi^{-1}(1-\frac{\alpha}{2})$.

To determine the anomalous regions, since the Gaussian kernel is not
localized, we threshold the solution to \eqref{eq: sparsekernel}
with a small threshold $w$ to ensure noises are not detected. Consequently,
anomalous regions are estimated by $1(\hat{a}>w)$, where $1(x)$
is the indicator function. In our study, we select $w=0.005\hat{s}$.
Furthermore, as the number of points in anomalous regions increases,
the corresponding kernel size should decrease accordingly. Therefore,
we update the bandwidth of kernel $K_{a}$ (i.e. $h_{a}$ ) proportionally
to the sampling resolution in anomalous regions. That is, $h_{a}=c_{h}\max_{r\in\hat{a}}\min_{r_{k}}\|r-r_{k}\|$.
From the simulation study, we found $c_{h}=0.2$ works reasonably
well.

\begin{algorithm}[tbp] 
\caption{APG algorithm for sparse kernel estimation of anomalies} 
\SetKwBlock{Initialize}{initialize}{end}
\Initialize{

Choose a basis for the background as $B$

$\theta_{a}^{(0)}=0$

}

\While{$\ensuremath{|\theta_{a}{}^{(k-1)}-\theta_{a}^{(k)}|>\epsilon}$}{

Update $\theta_{a}^{(k+1)}$ by $\theta_{a}^{(k+1)}=S_{\frac{\gamma}{2}}(x^{(k)}+K_{a}^{T}(e-K_{a}x^{(k)})))$

Update $t_{k+1}=\frac{1+\sqrt{1+4t_{k}^{2}}}{2}$

Update $x^{(k+1)}=\theta_{a}^{(k)}+\frac{t_{k}-1}{t_{k+1}}(\theta_{a}^{(k)}-\theta_{a}^{(k-1)})$

}

\label{alg: sparsekernel}

\end{algorithm}

\section{Simulation Study}

\begin{figure}
\subfloat[Simulated anomalies]{\includegraphics[width=0.3\linewidth]{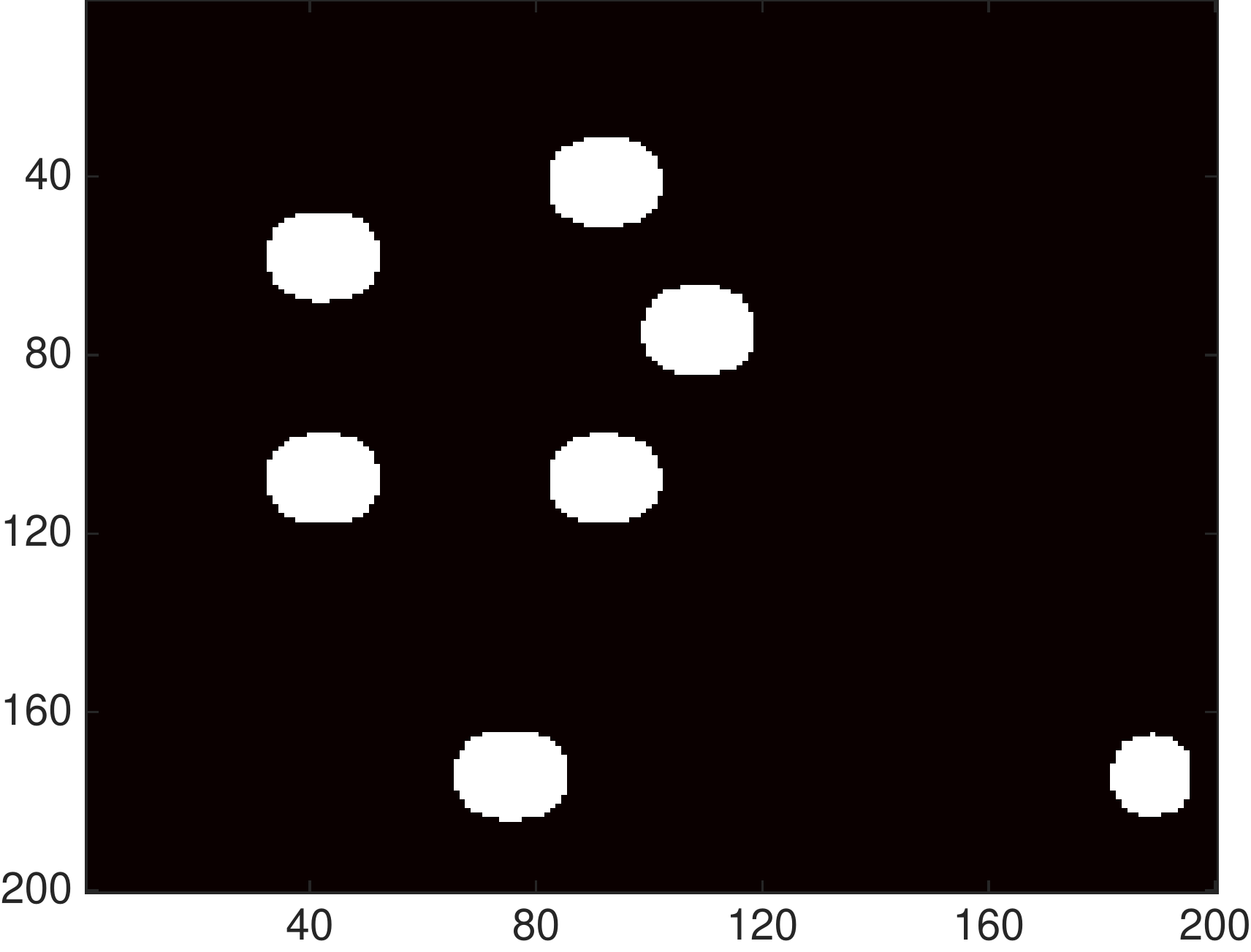}\label{Fig: defect-1}

}\hfill{}\subfloat[Simulated functional mean]{\includegraphics[width=0.3\linewidth]{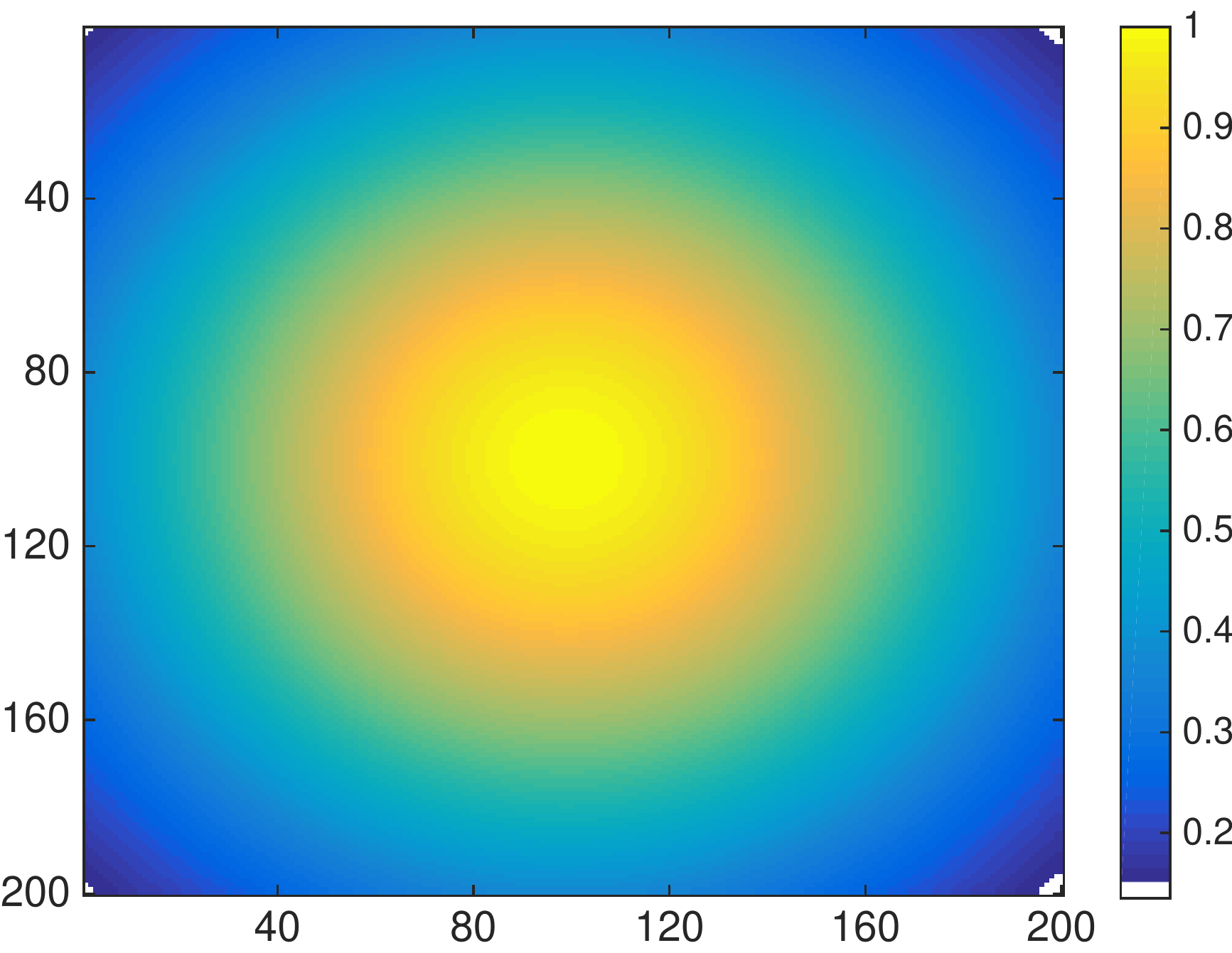}\label{Fig: heat}

}\hfill{}\subfloat[Simulated images coupled with noise and anomalies]{\includegraphics[width=0.3\linewidth]{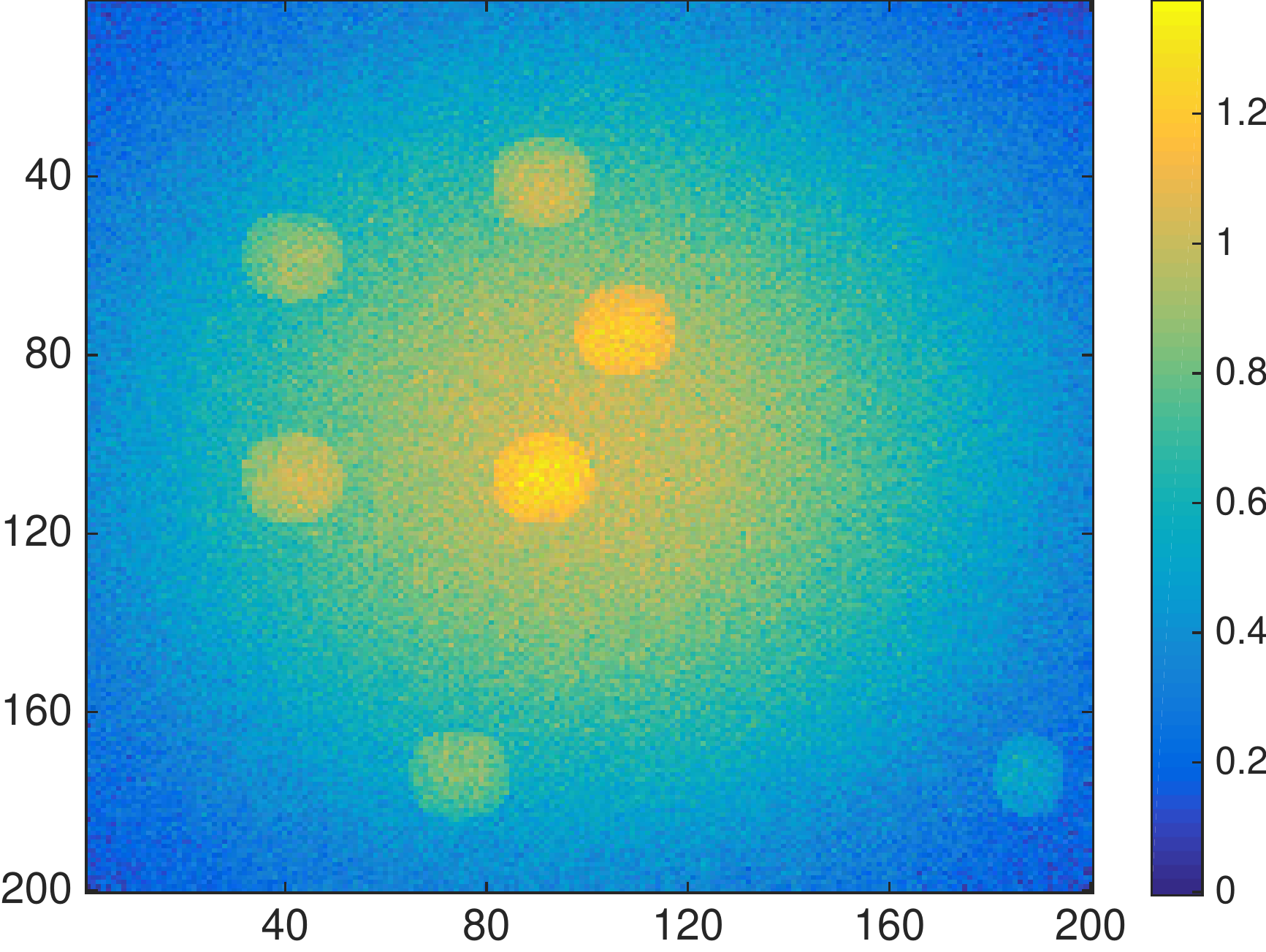}\label{Fig: Original}

}

\caption{Simulated images with both functional mean and anomalies}
\label{Fig: simulation}
\end{figure}

To evaluate the performance of the proposed methodology, we simulate
$200\times200$ images with a smooth functional mean denoted by matrix
$M$ whose elements are obtained by evaluating $M(x,y)=\exp(-\frac{(x^{2}+y^{2})}{4})$
at points $x=\frac{i}{201},y=\frac{j}{201};i,j=1,\cdots,200$. In
this study, $7$ anomaly clusters are generated by $A=B_{s}A_{s}B_{s}^{T}$,
in which $B_{s}$ is a cubic B-spline basis with 13 knots, and $A_{s}$
is a 13 by 13 sparse matrix with 7 randomly selected non-zero entries
denoted by $S_{A}$. The elements of $A_{s}$ are defined by $A_{s}(i,j)=\delta\cdot1(a_{ij}\in S_{A})$,
where $\delta=0.3$ characterizes the intensity difference between
anomalies and the functional mean. Random noises $E$ are generated
from $E\overset{}{\sim}NID(0,\sigma^{2})$ with $\sigma=0.05$. Finally,
the set of $200\times200$ simulated images, $Y$, is generated by
adding the anomalies and random noises to the functional mean, i.e.,
$Y=M+A+E$. A sample of simulated functional mean, anomalies, and
a noisy image with anomalies are shown in Figure \ref{Fig: simulation}.
The goal of this simulation study is to accurately estimate anomalous
regions with the least number of sampled points.

We compare our proposed adaptive sampling framework, $\mathrm{\mathbf{\mathrm{AKM^{2}D}}}$,
with four benchmark methods, the random sampling method (designated
by ``Random'') and multi-resolution grid sampling (designated by
``Grid''), adaptive maximum-minimum design (designated by ``DOE''),
and adaptive Gaussian process criterion (designated by ``Variance'').
In the Random sampling method, the sampled points are selected purely
at random. In Grid sampling, the sampled points are first selected
on a $15\times15$ coarse grid. If $p_{a}>0.5$, a finer grid with
a five-times-higher resolution is then used to sample within the coarse
grid containing anomalous points. For adaptive maximum-minimum design,
the point is selected by maximizing the minimum distance of the entire
sampling space. For adaptive Gaussian process criterion, we first
fit a Gaussian process model to the entire space. Second, the point
is selected at the point with the largest confidence interval of the
fitted value. We apply the proposed estimation method to the sampled
points obtained by both $\mathrm{\mathbf{\mathrm{AKM^{2}D}}}$ and
the benchmarks to estimate the anomalous regions. In this way, the
difference in anomaly detection performance can only be attributed
to the sampling strategy.

To evaluate the anomaly quantification accuracy, we propose to use
the precision, recall, and F1-score to evaluate the pixel-level image
segmentation accuracy. The average value and standard deviations of
the following criteria are computed over $5000$ simulation replications:
Precision, defined as the percentage of detected anomalies by the
algorithm that are indeed anomalous; Recall, defined as the percentage
of the true anomalous regions detected by the algorithm; F-measure,
defined as the harmonic mean of precision and recall; Exploitation
Ratio (ER), defined as the percentage ratio of sampled points in the
true anomalous regions to the total number of sampled points; Anomaly
Max-Min Distance (AMMD), defined as the maximum distance of points
in the true anomalous region to the nearest sampled point; Max-Min
Distance (MMD), defined as the maximum distance of points in the entire
sampling space to the nearest sampled point; and the computational
time of the sampling procedure for each sampled point. Here, Precision,
Recall, and F-measure are related to the accuracy of anomaly estimation,
which evaluate how the algorithms locate all anomalies (e.g., exploration)
and how well it quantifies each anomalous region (e.g., exploitation).
AMMD, ER, and MMD are direct quantification on the performance of
the sampling algorithm in terms of exploration and exploitation. For
example, AMMD and ER are related to the exploitation performance of
the proposed $\mathrm{\mathbf{\mathrm{AKM^{2}D}}}$ algorithm. MMD
is related to the exploration of the proposed $\mathrm{\mathbf{\mathrm{AKM^{2}D}}}$
algorithm.

These average values and standard deviations of 250 points and 400
points are reported in Table 1 and Table 2, respectively. From these
tables, it is clear that the proposed $\mathrm{\mathbf{\mathrm{AKM^{2}D}}}$
overall outperforms other benchmark methods. For example, with $250$
sampled points, the recall of $\mathrm{\mathbf{\mathrm{AKM^{2}D}}}$
is $0.78(0.0021)$ indicating that $78\%$ (with the standard deviation
0.21\%) of the anomalous regions have been detected by $\mathrm{\mathbf{\mathrm{AKM^{2}D}}}$
with only $250$ points. This is much higher than the recall of benchmarks
that is at most about $27\%$. Although benchmark methods have slightly
higher precision, the overall classification accuracy, measured by
F is in favor of $\mathrm{\mathbf{\mathrm{AKM^{2}D}}}$. The F-measure
of $\mathrm{\mathbf{\mathrm{AKM^{2}D}}}$ is around $0.72$ (with
standard deviation $0.0011$), while it is at most $0.39$ for benchmark
methods. The MMD value of the $\mathrm{\mathbf{\mathrm{AKM^{2}D}}}$
is only larger than the pure DOE method and smaller than all other
methods. This is expected since DOE only focus on the exploration
of the entire sampling space without paying attention to any focus
sampling. Therefore, the AMMD values of the $\mathrm{\mathbf{\mathrm{AKM^{2}D}}}$
are also much smaller than the benchmark methods, which indicates
the proposed $\mathrm{\mathbf{\mathrm{AKM^{2}D}}}$ achieves better-focused
sampling near the anomalous regions.

Similarly, the ER of $\mathrm{\mathbf{\mathrm{AKM^{2}D}}}$ with $250$
sampled points method is around $18\%$ (with standard deviation $3\%$),
$3.6$ times larger than that of Random, Grid, DOE (around $5\%$)
and Variance (around 2\%). This implies that the proposed method is
able to quickly locate anomalous regions and sample about $3.6$ times
more points in those regions than benchmark methods. Note that the
area of anomalous regions covers about $5.4\%$ of the entire sampling
space. However, $\mathrm{\mathbf{\mathrm{AKM^{2}D}}}$ with around
$0.6\%$ of the full sampled points ($250$ sampled points out of
$200\times200$), is able to detect at least $78\%$ of the true anomalous
regions. If we increase the number of sampled points to $400$, this
number increases to $88\%$, whereas for other benchmark methods it
is at most $65\%$. The main reason for the poor performance of Random,
DOE, and Variance is that they lack any abilities to focus on the
discovered anomalous regions. Grid has some power of focusing on the
discovered anomalous regions. The reason for the bad performance of
the Variance method lies in the stability and boundary issue. We observe
that the boundary is not correctly estimated with the Gaussian process
(GP) and the algorithm may tend to put more points in the boundary.
However, the fine sampling grid is rigid, and hence it is not flexible
to detect arbitrarily shaped anomalies. Although $\mathrm{\mathbf{\mathrm{AKM^{2}D}}}$
is slightly slower than the benchmarks, all methods except Variance
satisfy the real-time speed requirement for online sensing. Finally,
the standard deviation of the proposed method in all these criteria
is also quite small, which implies the proposed methods are robust
to random anomaly locations and random noises.

\begin{figure}
\subfloat[MMD]{\includegraphics[width=0.45\linewidth]{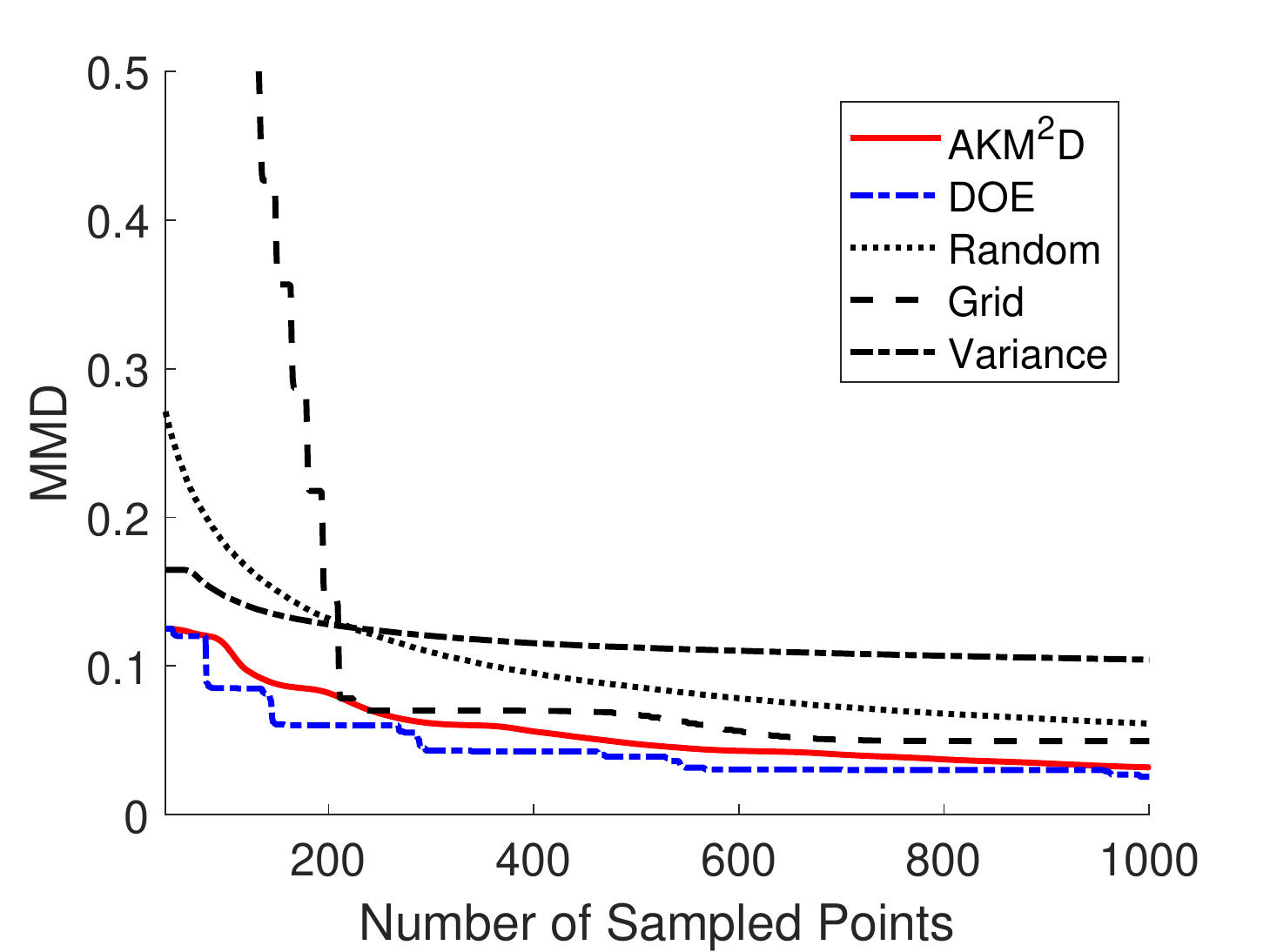}

}\hfill{}\subfloat[AMMD]{\includegraphics[width=0.45\linewidth]{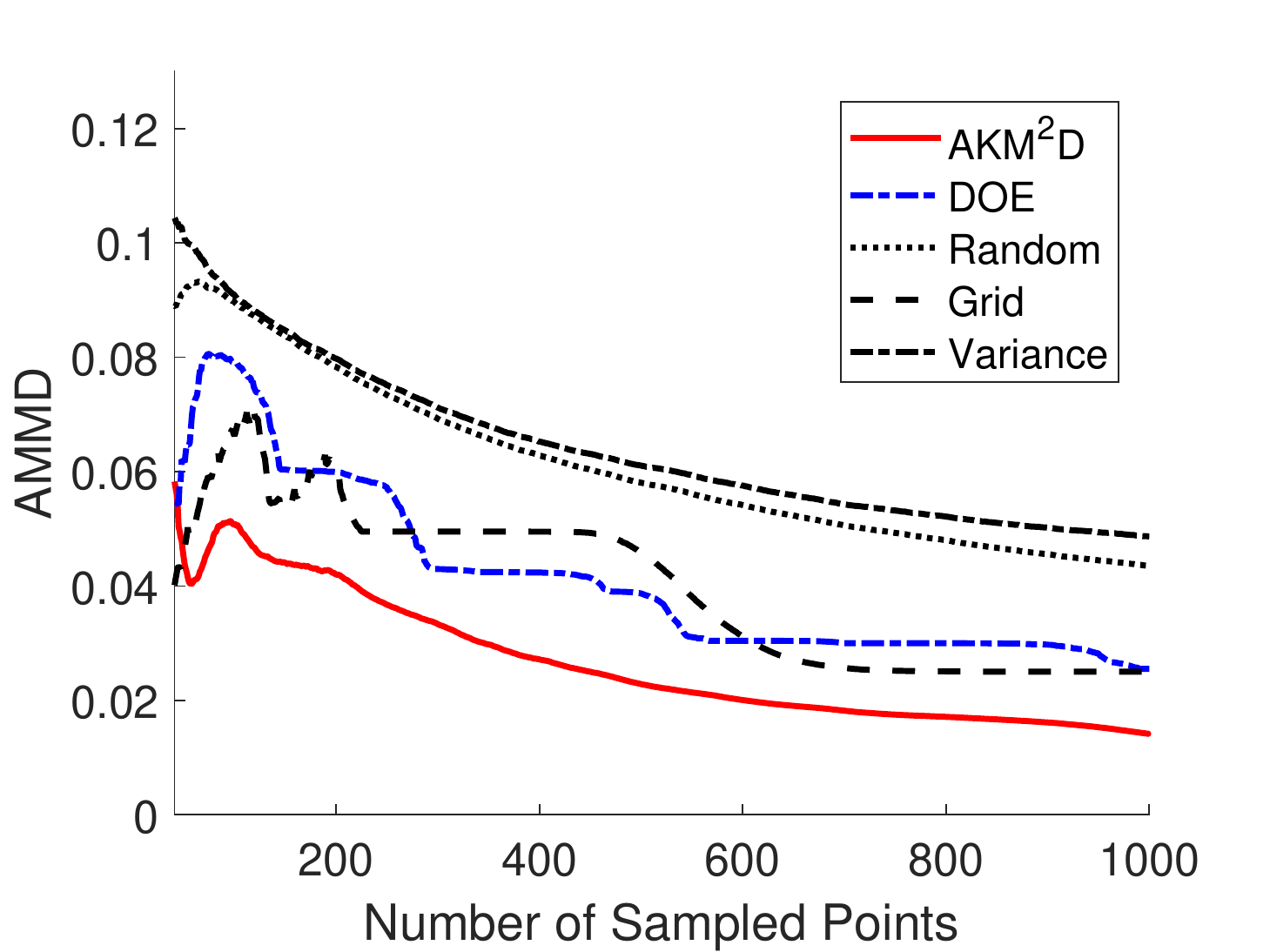}

}

\subfloat[F-measure]{\includegraphics[width=0.45\linewidth]{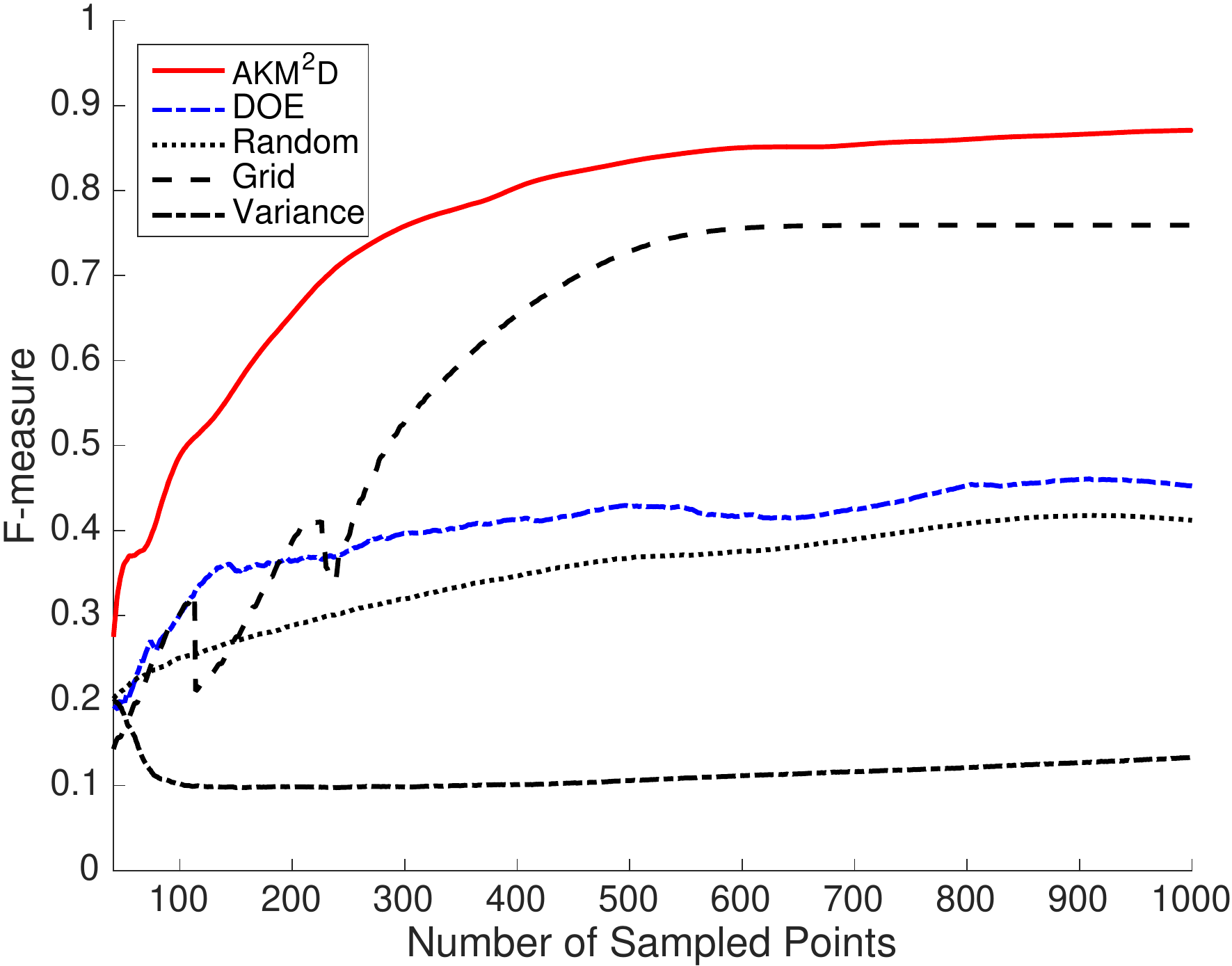}

}\hfill{}\subfloat[Exploration ratio]{\includegraphics[width=0.45\linewidth]{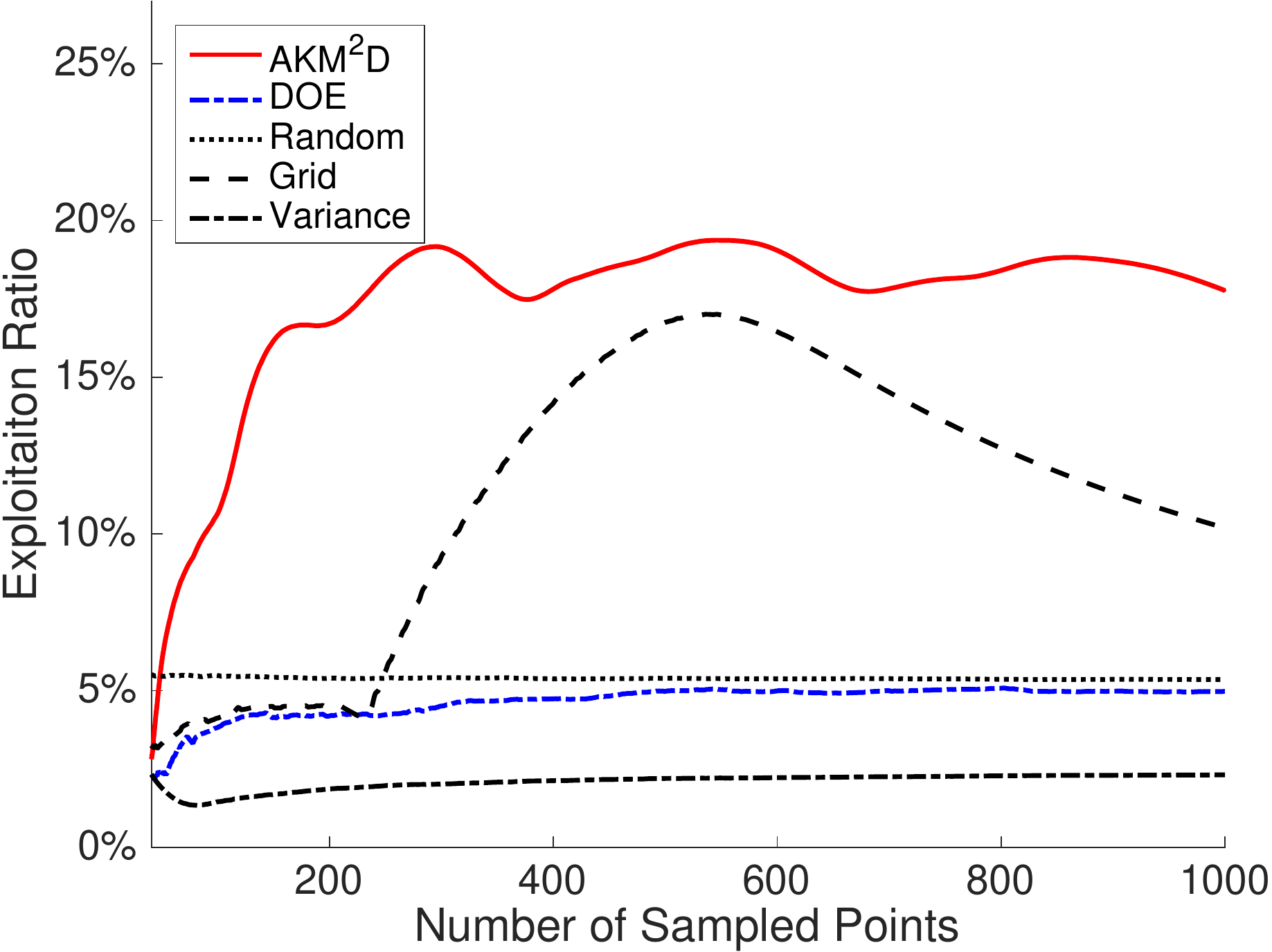}

}

\caption{MMD, AMMD, F-measure and Exploitation Ratio}
\label{Fig: F}
\end{figure}

The average values of the MMD, AMMD, F-measure, and the ER against
the iteration number (number of sampled points) are also plotted in
Figure \ref{Fig: F}. From this figure, we can conclude that the F-measure
of $\mathrm{\mathbf{\mathrm{AKM^{2}D}}}$ is strictly better than
other benchmark methods for any number of sampled points. Furthermore,
the ER of $\mathrm{\mathbf{\mathrm{AKM^{2}D}}}$ increases to $18\%$
with only $200$ points and then oscillating around $18\%$, showing
its superiority to quickly locate and sample the anomalous regions.
The MMD of the proposed $\mathrm{\mathbf{\mathrm{AKM^{2}D}}}$ is
better than Random, Grid, Variance, and only second to DOE. However,
the AMMD of the proposed $\mathrm{\mathbf{\mathrm{AKM^{2}D}}}$ is
much better than all other benchmark methods, which demonstrates the
supreme overall sampling performance. The ER of Grid stays at $4\%$
during the coarse grid sampling and only begin to increase up to $16\%$
when performing the fine-grid sampling (after $225$ points). Finally,
the ER of Random, DOE, and Variance stays lower than $6\%$, which
is the percentage of true anomalous regions.

\begin{table}
\caption{Anomaly Detection Result with $250$ sampled points}

\centering

\scriptsize%
\begin{tabular}{|c|ccccc|}
\hline 
\multirow{2}{*}{Methods} & \multicolumn{5}{c|}{$250$ sampled points}\tabularnewline
\cline{2-6} \cline{3-6} \cline{4-6} \cline{5-6} \cline{6-6} 
 & $\mathrm{\mathbf{\mathrm{AKM^{2}D}}}$ & Random & Grid & Variance & DOE\tabularnewline
\hline 
Precision & $0.69(0.0011)$ & $0.80(0.0023)$ & $0.80(0.0013)$ & $0.74(0.006)$ & $0.74(0.0033)$\tabularnewline
Recall & $0.78(0.0021)$ & $0.19(0.0021)$ & $0.27(0.0015)$ & $0.05(0.0008)$ & $0.26(0.0017)$\tabularnewline
F & $0.72(0.0011)$ & $0.30(0.0027)$ & $0.39(0.0016)$ & $0.098(0.0014)$ & $0.38(0.0021)$\tabularnewline
ER & $18\%
$ & $5.4\%(0.03\%)$ & $5.6\%(0.03\%)$ & $1.9\%(0.03\%)$ & $4.2\%(0.03\%)$\tabularnewline
AMMD & $0.036(0.002)$ & $0.073(0.00)$ & $0.049(0.00)$ & $0.07(0.00)$ & $0.057(0.0002)$\tabularnewline
MMD & $0.068(0.0001)$ & $0.119(0.00)$ & $0.070(0.00)$ & $0.122(0.0002)$ & $\mathbf{0.060}(0.00)$\tabularnewline
Time & $0.0046$s & $0.0026$s & $\mathbf{0.0025}$\textbf{s} & $0.473$s & $0.003s$\tabularnewline
\hline 
\end{tabular}

\label{table: anomalytable}
\end{table}

\begin{table}
\caption{Anomaly Detection Result with $400$ sampled points}

\centering

\scriptsize

\begin{tabular}{|c|ccccc|}
\hline 
\multirow{2}{*}{Methods} & \multicolumn{5}{c|}{$400$ sampled points}\tabularnewline
\cline{2-6} \cline{3-6} \cline{4-6} \cline{5-6} \cline{6-6} 
 & $\mathrm{\mathbf{\mathrm{AKM^{2}D}}}$ & Random & Grid & Variance & DOE\tabularnewline
\hline 
Precision & $0.75(0.009)$ & $\mathbf{0.80}(0.002)$ & $0.66(0.001)$ & $0.68(0.005)$ & $0.76(0.0025)$\tabularnewline
Recall & $\mathbf{0.88}(0.0015)$ & $0.23(0.0019)$ & $0.65(0.0014)$ & $0.05(0.0008)$ & $0.28(0.0011)$\tabularnewline
F & $\mathbf{0.80}(0.0003)$ & $0.35(0.0023)$ & $0.65(0.0005)$ & $0.101(0.0014)$ & $0.41(0.0015)$\tabularnewline
ER & $\mathbf{18\%}(0.03\%)$ & $5.4\%(0.03\%)$ & $14.17\%(0.05\%)$ & $2.1\%(0.04\%)$ & $4.7\%(0.03\%)$\tabularnewline
AMMD & $\mathbf{0.027}(0.001)$ & $0.063(0.0002)$ & $0.049(0.00)$ & $0.07(0.00)$ & $0.04(0.00)$\tabularnewline
MMD & $0.056(0.00)$ & $0.095(0.0003)$ & $0.070(0.00)$ & $0.11(0.0002)$ & $\mathbf{0.04}(0.00)$\tabularnewline
Time & $0.0046$s & $0.0026$s & $\mathbf{0.0025}$\textbf{s} & $0.473$s & $0.003s$\tabularnewline
\hline 
\end{tabular}

\label{table: anomalytable-1}
\end{table}

\begin{figure}
\subfloat[250 sampled points]{\includegraphics[width=0.45\linewidth]{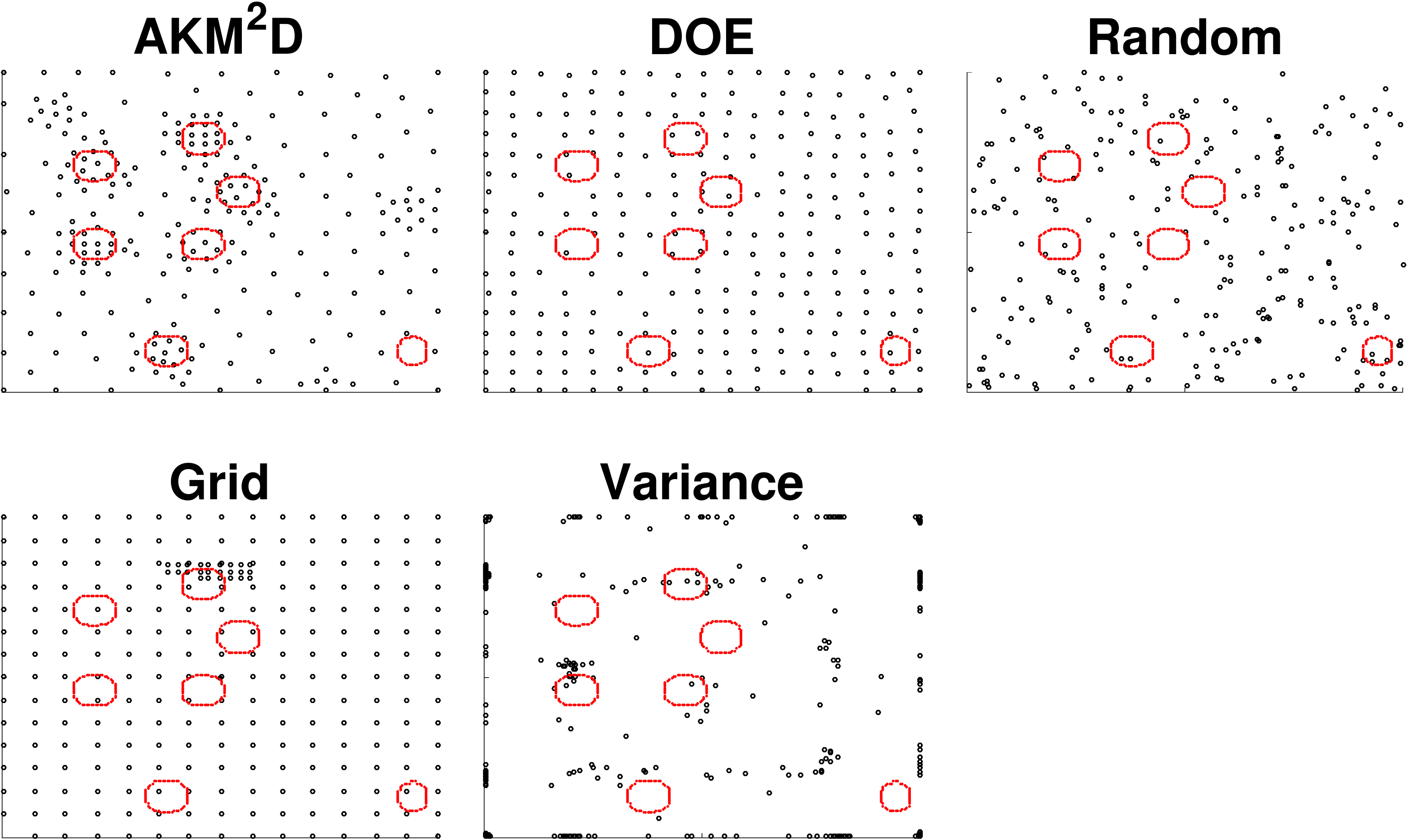}\label{Fig: 250}

}\hfill{}\subfloat[400 sampled points]{\includegraphics[width=0.45\linewidth]{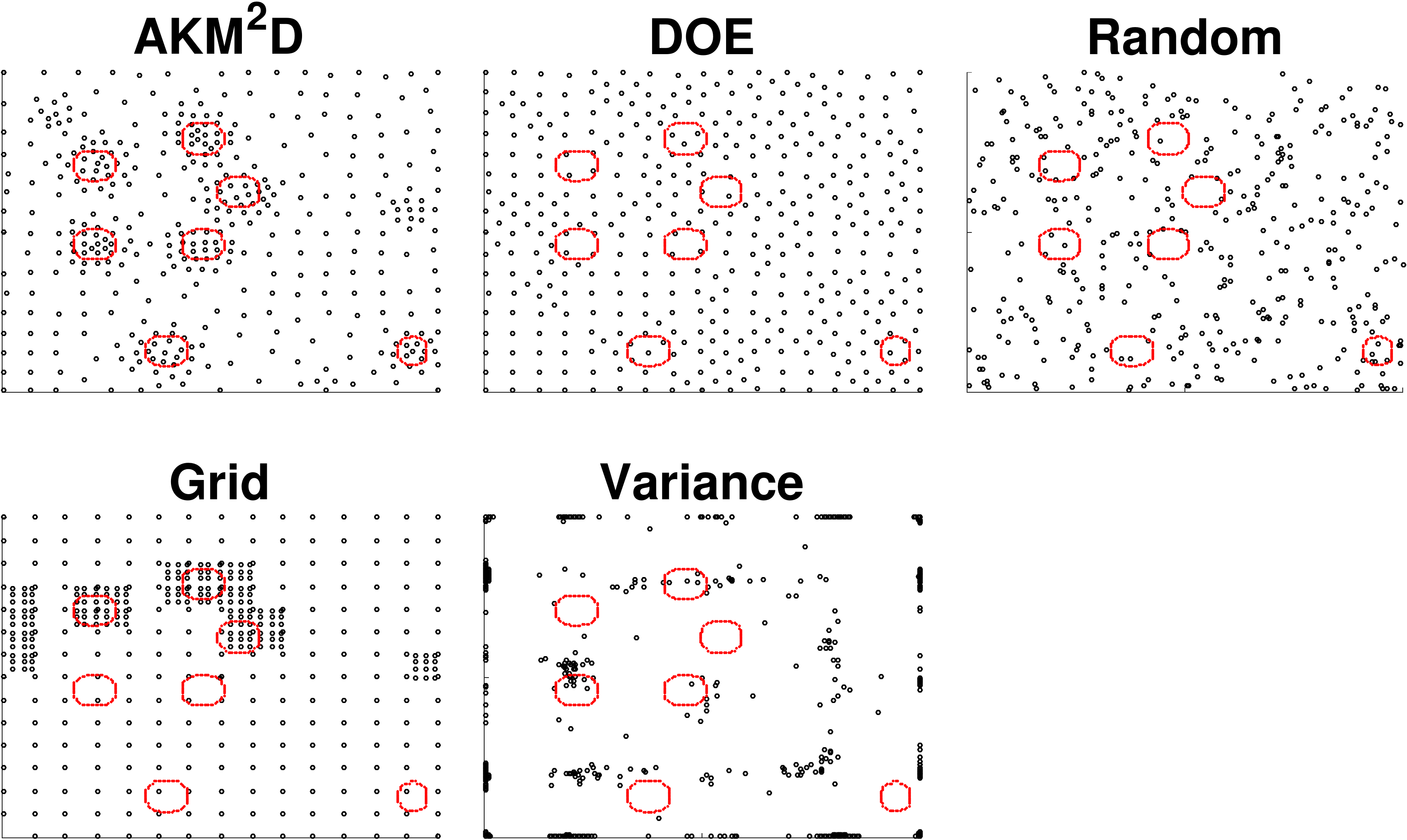}\label{Fig: 400}

}

\caption{Sampled point pattern for all methods for 250 and 400 points}

\label{Fig: pointpattern}
\end{figure}

\begin{figure}
\subfloat[250 sampled points]{\includegraphics[width=0.45\linewidth]{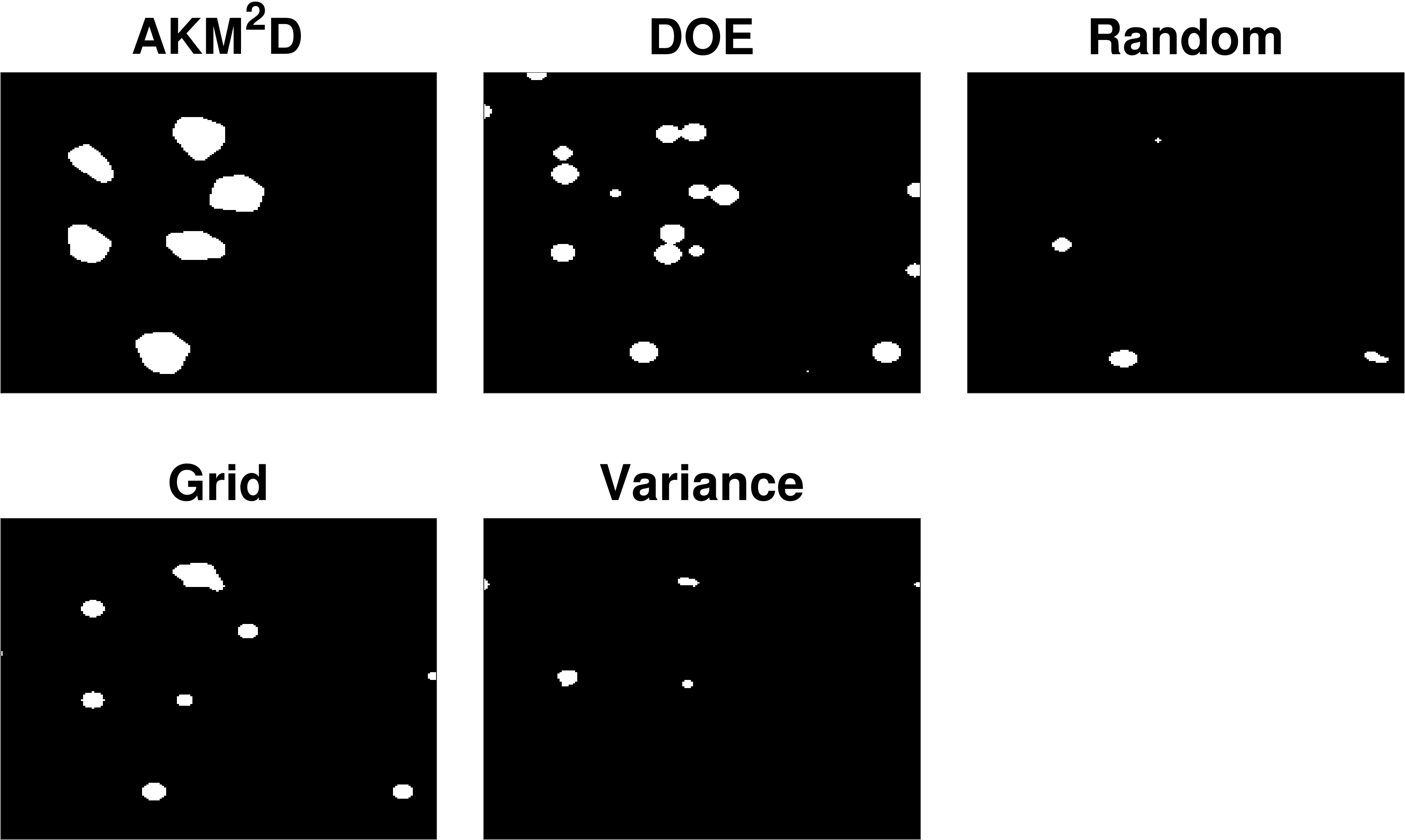}\label{Fig: 250anomaly}

}\hfill{}\subfloat[400 sampled points]{\includegraphics[width=0.45\linewidth]{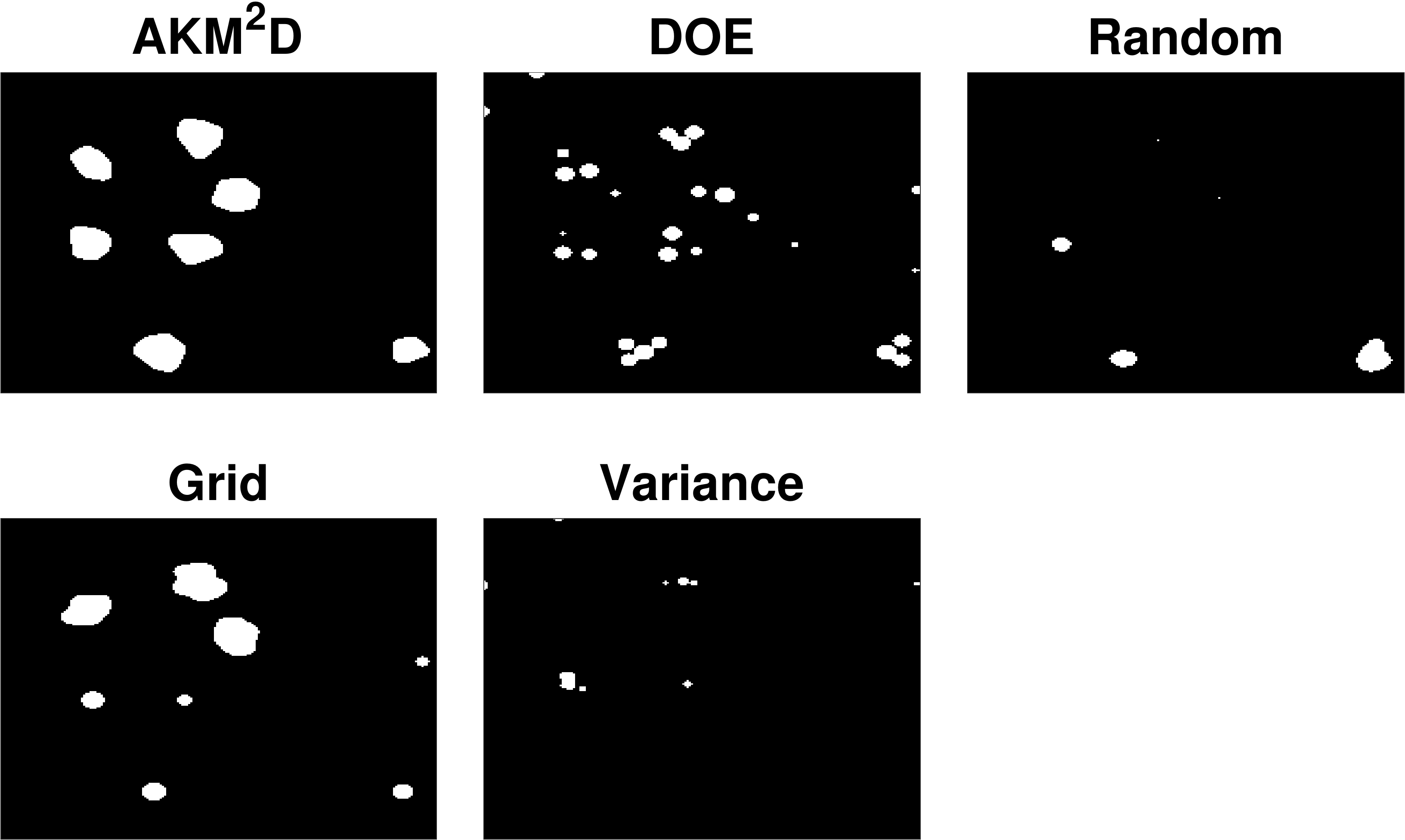}\label{Fig: 400anomaly}

}

\caption{Anomaly estimation result for all methods for 250 and 400 points}
\label{Fig: simanomaly}
\end{figure}

Furthermore, we investigate the pattern of sampled points (with $250$
and $400$ points) in Figure \ref{Fig: pointpattern}. From the figure,
we can observe that with only 250 sampled points, $\mathrm{\mathbf{\mathrm{AKM^{2}D}}}$
discovers all anomalous regions but one, with a better space-filling
point distribution. However, Random, Variance, and DOE only put a
few points on the anomalous regions, which fail to detect any of the
anomalous regions and Grid can only detect one. On the other hand,
400 sampled points are enough for $\mathrm{\mathbf{\mathrm{AKM^{2}D}}}$
to detect all 7 anomalous regions. However, again Random, Variance,
and DOE fails to discover any anomalous regions and Grid finishes
with the fine-grid sampling of only three regions. Also, we plot the
detected anomalies corresponding to $250$ and $400$ sampled points
in Figure \ref{Fig: simanomaly}, which again indicates the superior
performance of $\mathrm{\mathbf{\mathrm{AKM^{2}D}}}$ in anomaly detection.

We also plot different sampling point patterns with different tuning
parameters $\lambda,u,h$ in Figure 7-9. We can conclude that smaller
$\lambda$ and $u$ or larger $h$ tends to lead to a better exploration
of the entire background. Larger $\lambda$ and $u$ or smaller $h$
tends to lead to better exploitation of the anomaly. Therefore, the
balance of exploration and exploitation for different tasks will decide
what tuning parameter that we will use in the algorithm. Furthermore,
to help practitioners to understand what best tuning parameter combination
is suitable for their practical need, we also perform a complete tuning
sensitivity analysis in Appendix E.

\begin{figure}
\includegraphics[width=1\linewidth]{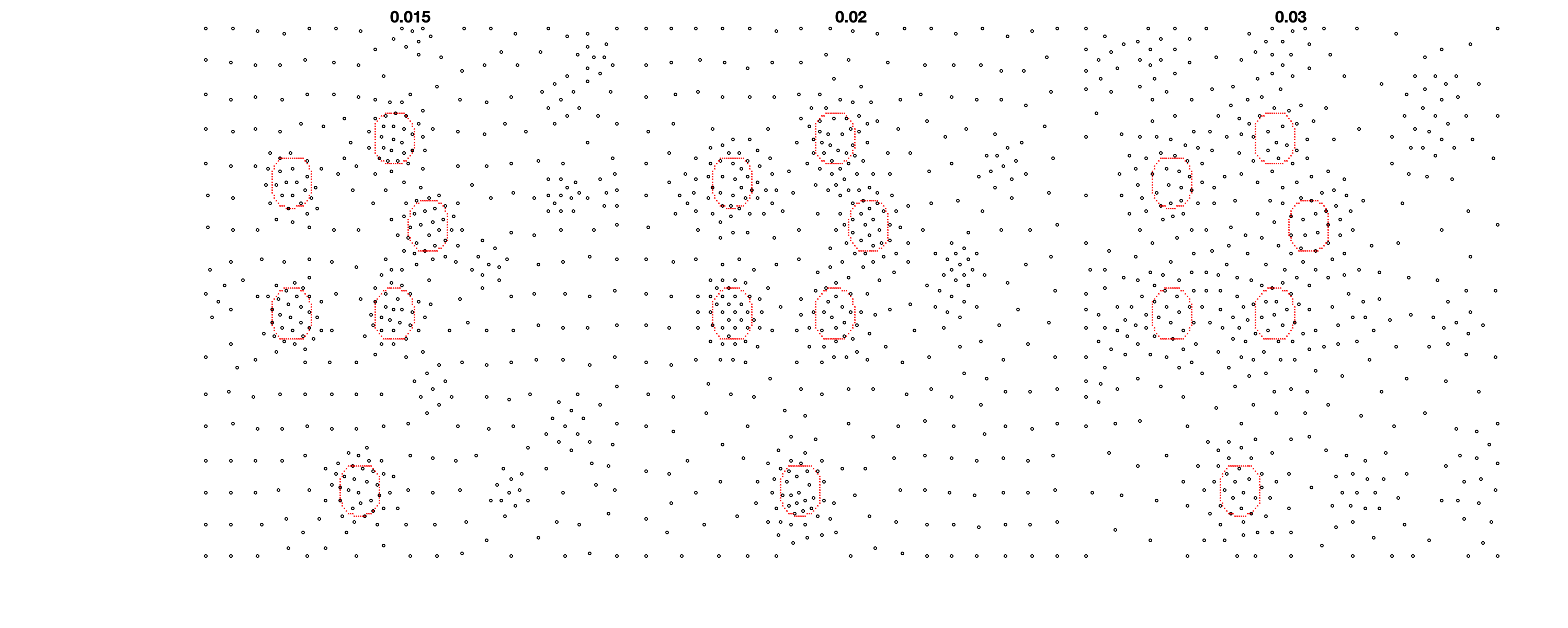}

\caption{Effect of $h$ (e.g., $h=0.015,0.02,0.03$ from left to right)}

\label{Fig: effecth}
\end{figure}

\begin{figure}
\includegraphics[width=1\linewidth]{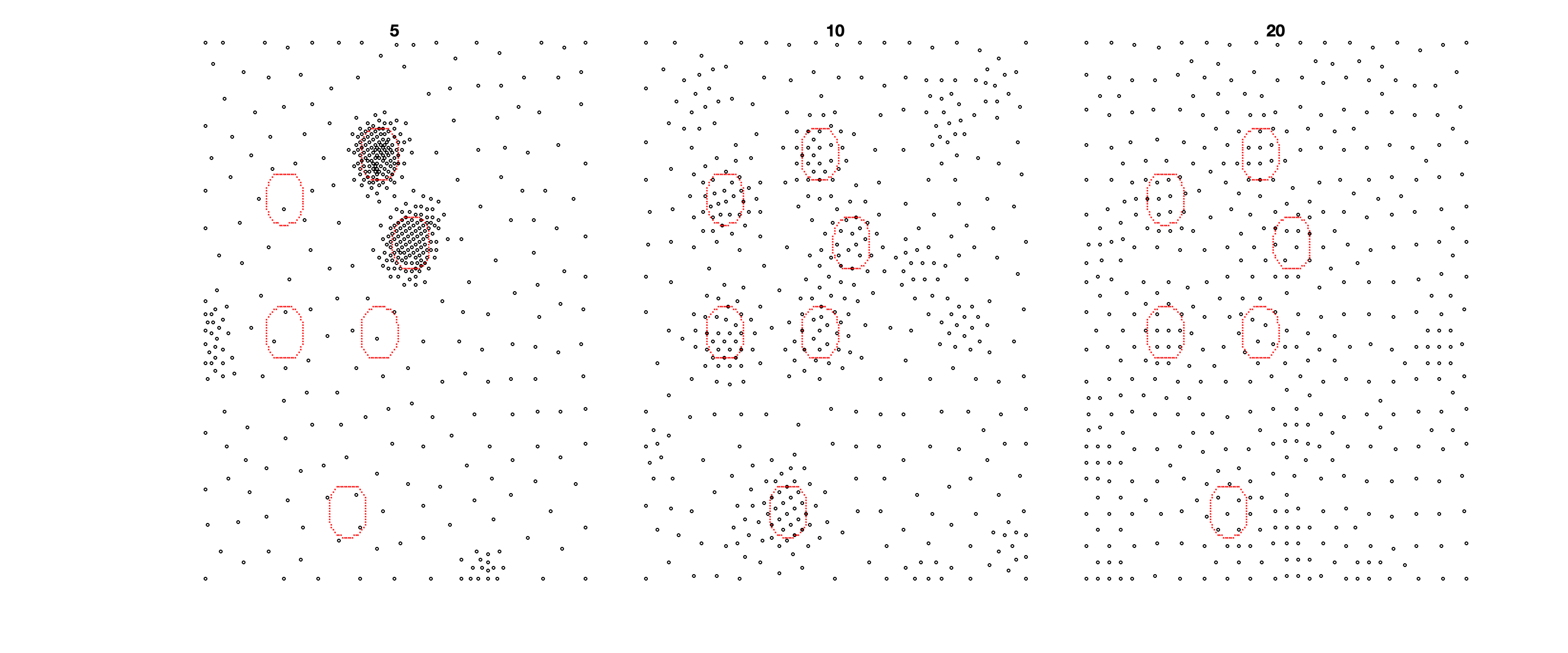}

\caption{Effect of $\lambda$ (e.g., $\lambda=5,10,20$ from left to right)}

\label{Fig: effectlambda}
\end{figure}

\begin{figure}
\includegraphics[width=1\linewidth]{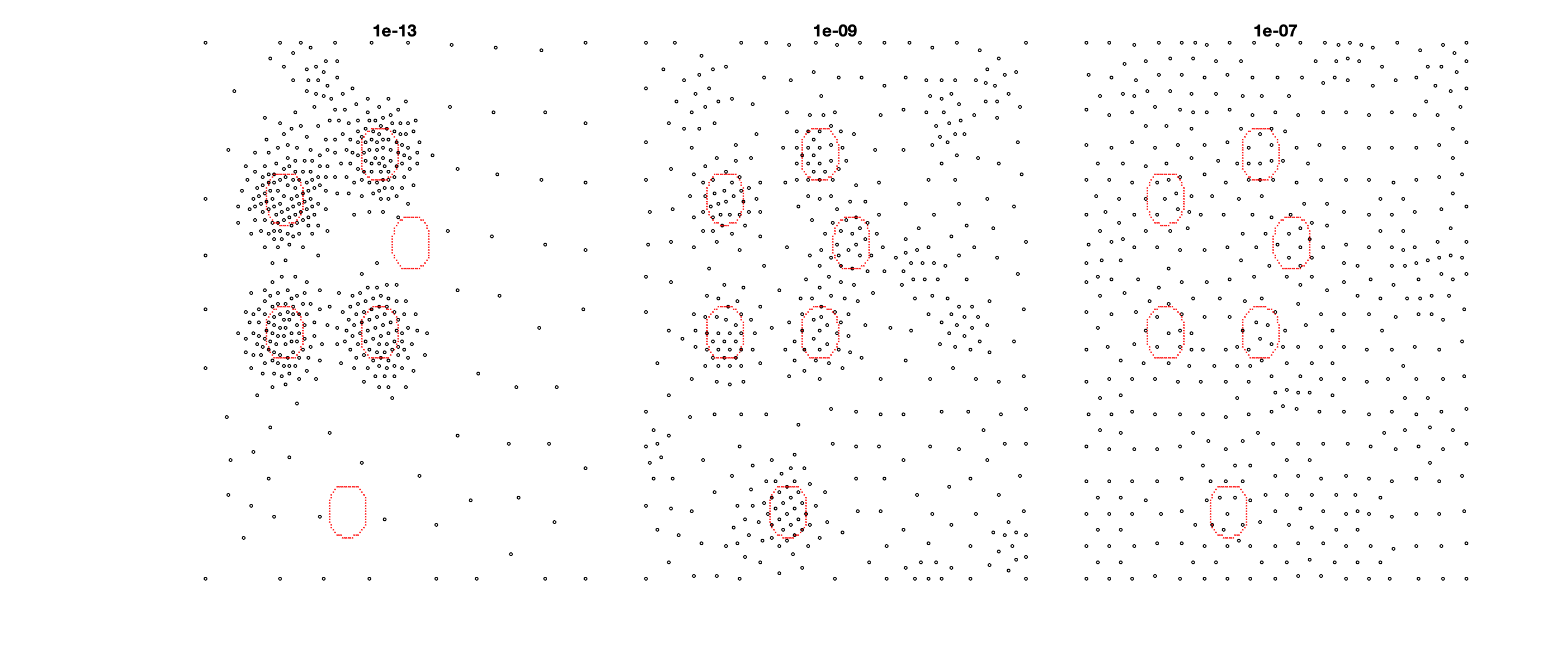}

\caption{Effect of $u$ (e.g., $u=10^{-13},10^{-9},10^{-7},$ from left to
right)}

\label{Fig: effectu}
\end{figure}

\section{Case Study \label{sec:Case-study}}

In this section, the proposed adaptive sampling and estimation framework
is applied to a real dataset in the NDE area. The case study pertains
to anomaly detection in composite laminates using a guided wave-field
(GW) inspection system. Lamb wave-based inspection is one of the popular
methods in NDE and structural health monitoring due to its high sensitivity
to detecting anomalies invisible to the naked eye \citep{mesnil2016sparse}.
However, existing GW techniques are point-based and require the whole-field
inspection of a specimen. The whole-field inspection is typically
a time-consuming process as it requires sensing of a large number
of points to avoid spatial aliasing and to achieve the desired resolution
\citep{mesnil2016sparse}. Therefore, it is vital to reduce the data
acquisition time by reducing the number of sampled points using an
adaptive sampling strategy.

\begin{figure}
\centering\includegraphics[width=0.8\linewidth]{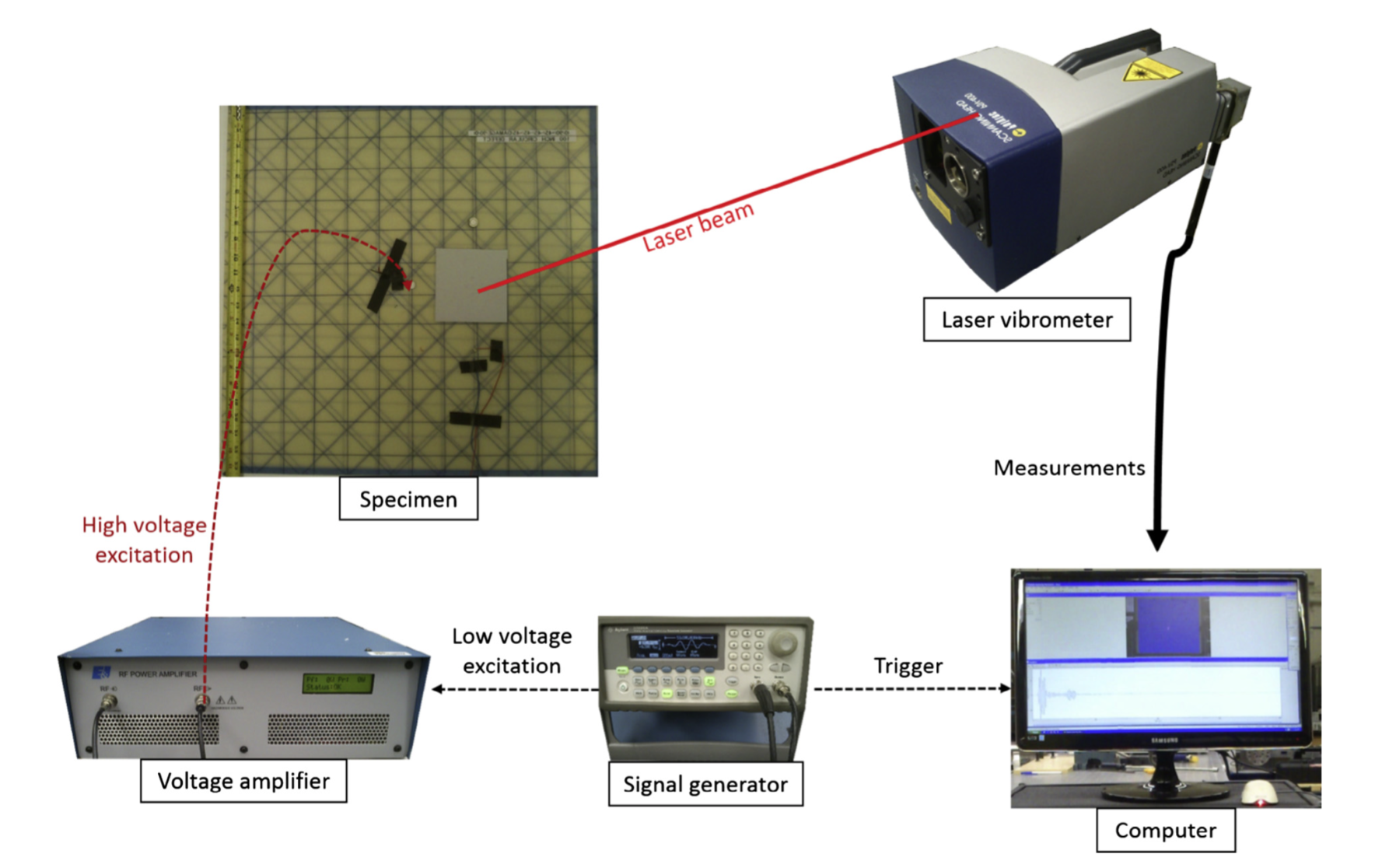}

\caption{Guided wavefield experiment setup \citep{mesnil2016sparse}}

\label{Fig: ExperimentSetup}
\end{figure}

\begin{figure}
\subfloat[Energy map of the entire wavefield]{\includegraphics[width=0.45\linewidth]{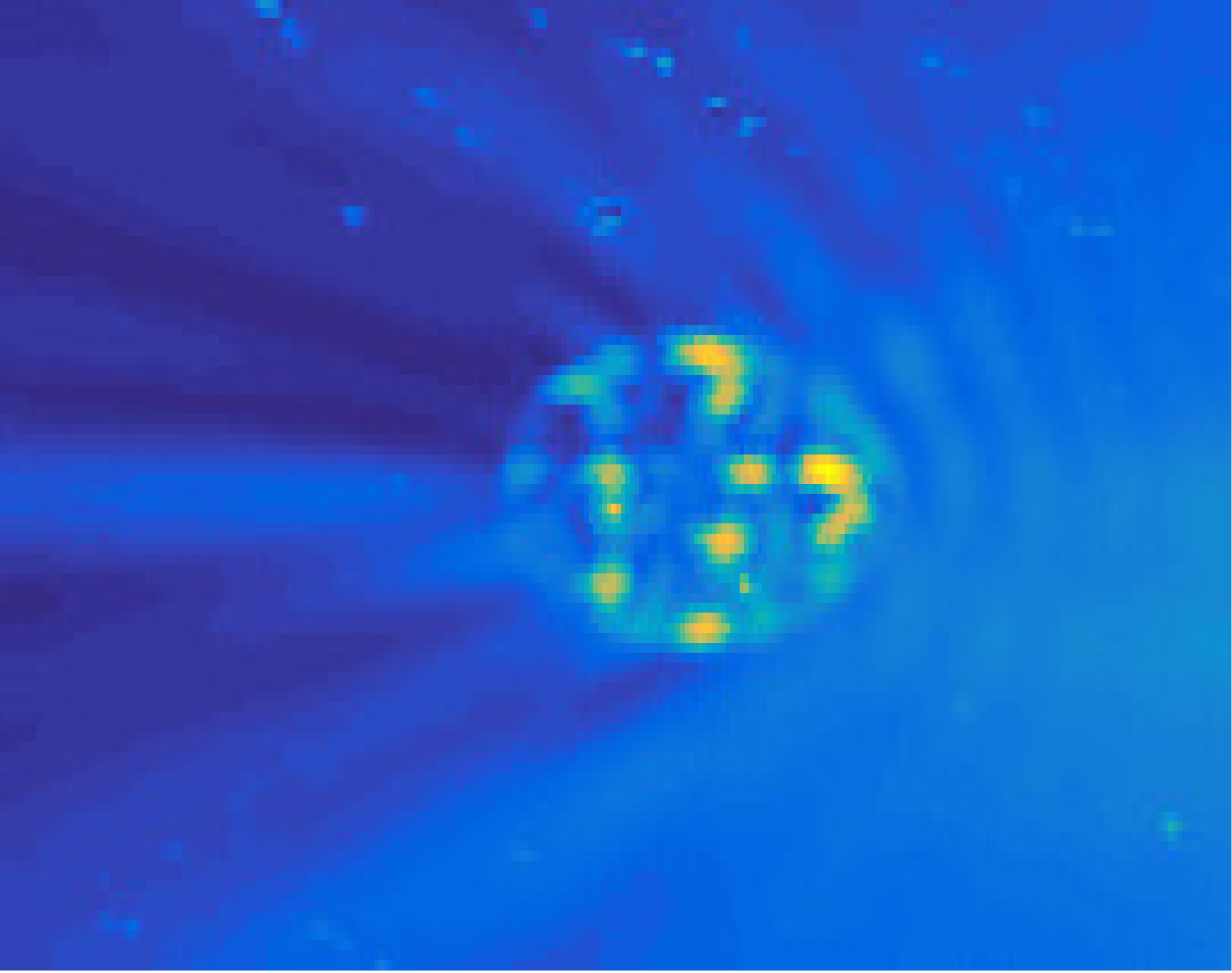}

\label{Fig: caseY}}\hfill{}\subfloat[Detected anomaly]{\includegraphics[width=0.45\linewidth]{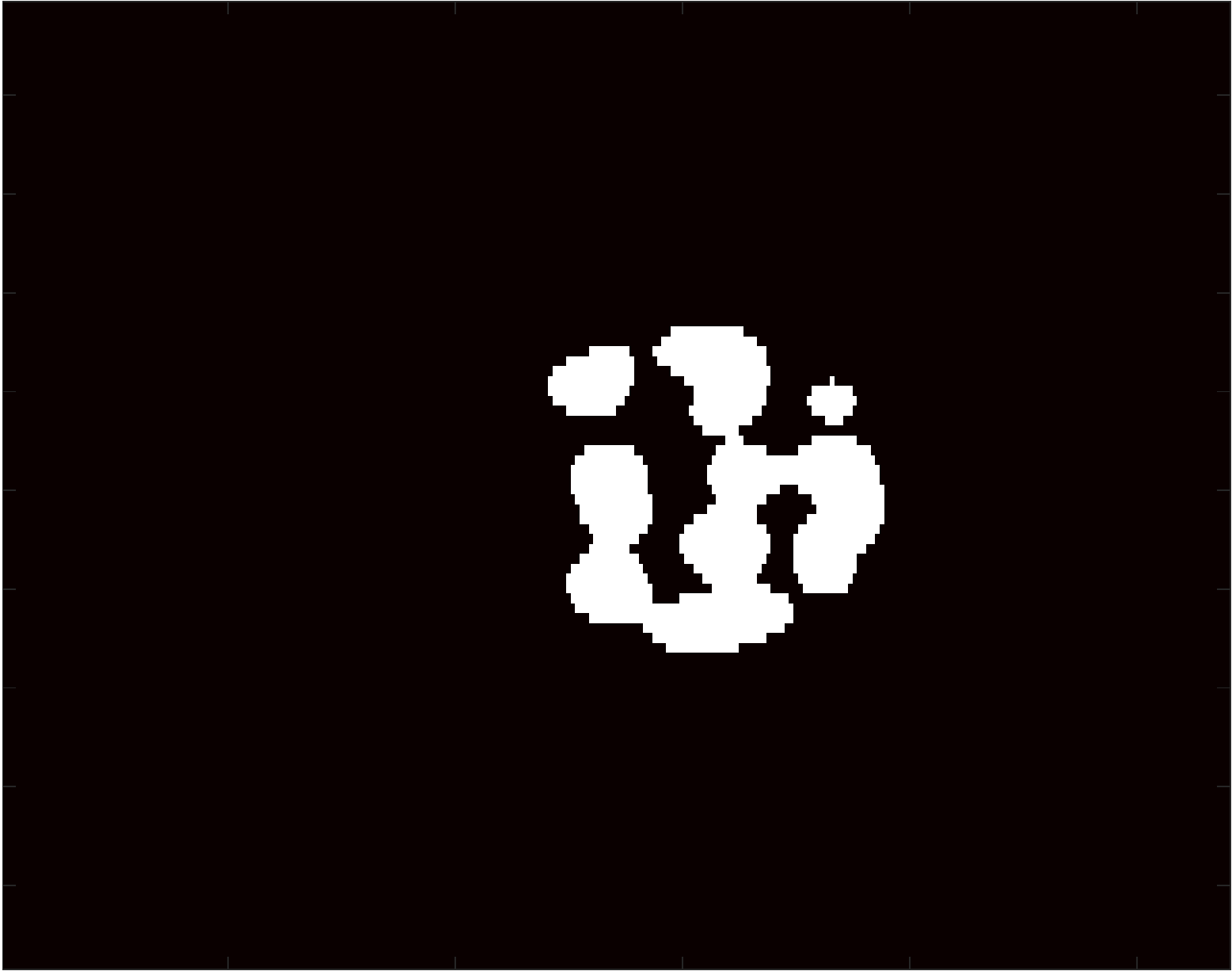}

\label{Fig: caseStrue}}\caption{Energy map of the entire wavefield and detected anomaly}
\end{figure}

The setup of our GW experiment is shown in Figure \ref{Fig: ExperimentSetup}.
A scanning laser Doppler vibrometer (SLDV) is employed for wavefield
measurement over a grid of points with the resolution of $270\times100$.
It takes around $4$ hours to inspect a $600\times600\times1.6$ mm
composite laminate with $8$ layers. The specimen contains several
artificial delaminations in the center, as shown in Figure \ref{Fig: caseY},
which is the energy map of the entire wavefield based on complete
sampling. To speed up the GW test so that it can be used for online
inspection, we reduce the number of sensing points by using adaptive
sampling strategies. For comparison purposes, we show detected anomalies
using a complete sampling strategy (i.e., Figure \ref{Fig: caseY})
in Figure \ref{Fig: caseStrue}. The objective is to achieve a similar
detection accuracy with the least number of sampled points.

\begin{figure}
\subfloat[F-measure]{\includegraphics[width=0.45\linewidth]{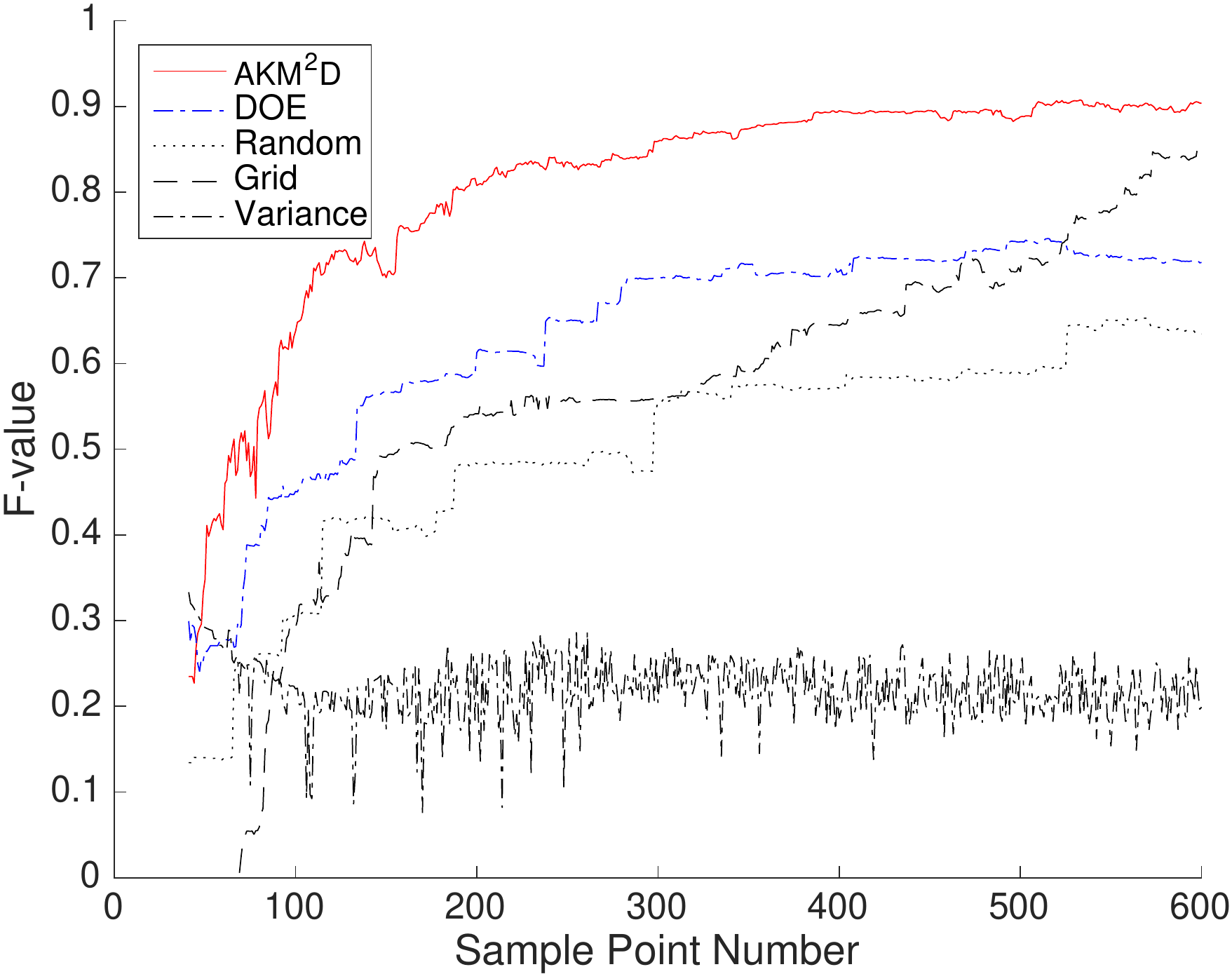}

}\hfill{}\subfloat[Exploitation Ration]{\includegraphics[width=0.45\linewidth]{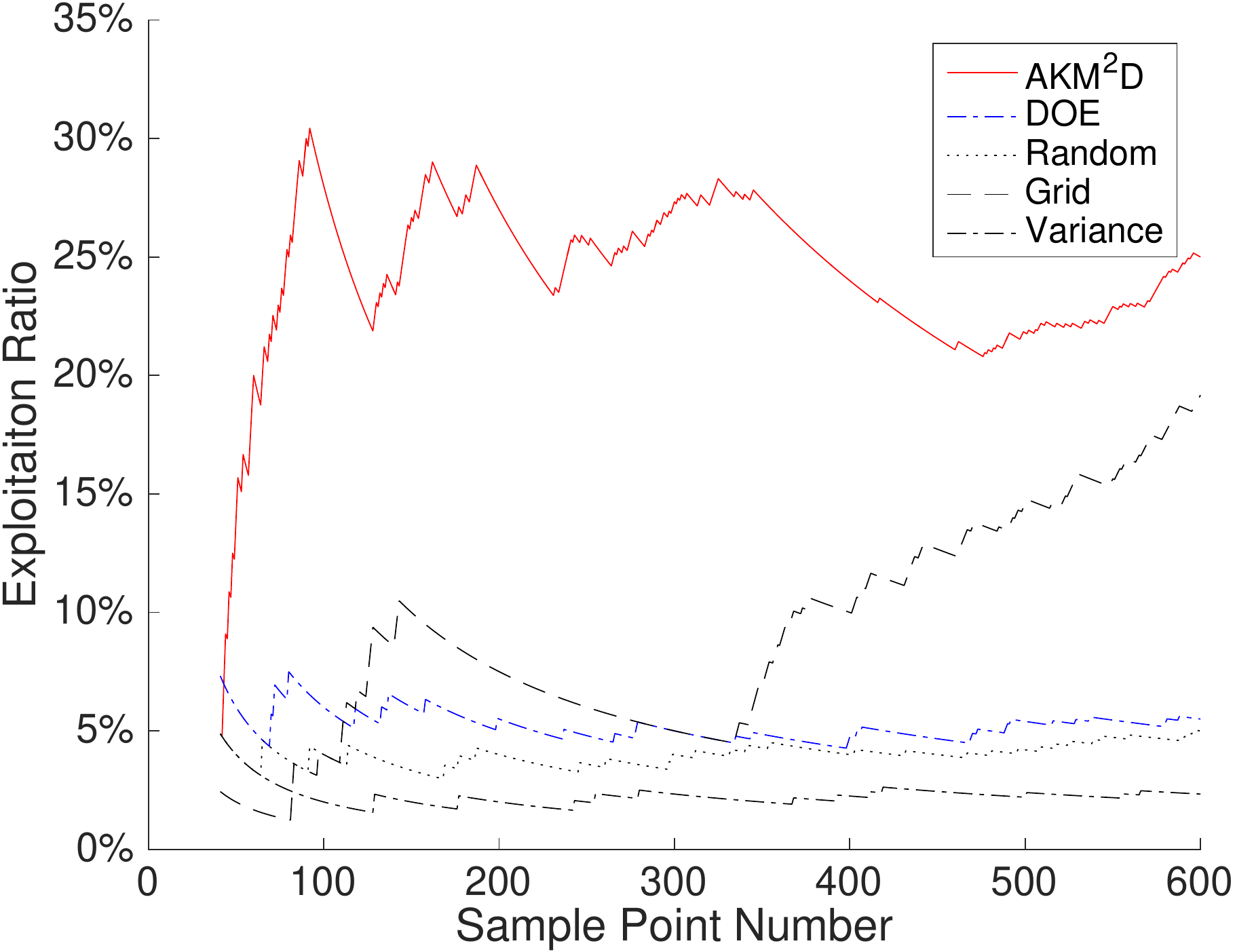}

}

\caption{F-measure and Exploitation Ration}

\label{Fig: caseF}
\end{figure}

We apply $\mathrm{\mathbf{\mathrm{AKM^{2}D}}}$ as well as four other
benchmark methods (i.e., Random, Grid, Variance, and DOE) for adaptive
sampling and use the proposed estimation methods for anomaly detection.
We compare the detection results obtained from the adaptive sampling
methods with those of the complete sampling, shown in Figure \ref{Fig: caseStrue},
(as the ground truth), and compute the F-measure and ER profiles depicted
in Figure \ref{Fig: caseF}. We can observe that with only $300$
points (1.1\% of complete sensing) $\mathrm{\mathbf{\mathrm{AKM^{2}D}}}$
is able to achieve the F-measure of $0.8$ much higher than those
of other benchmark methods, which is at most $0.5$.

\begin{figure}
\subfloat[200 sampled points]{\includegraphics[width=0.45\linewidth]{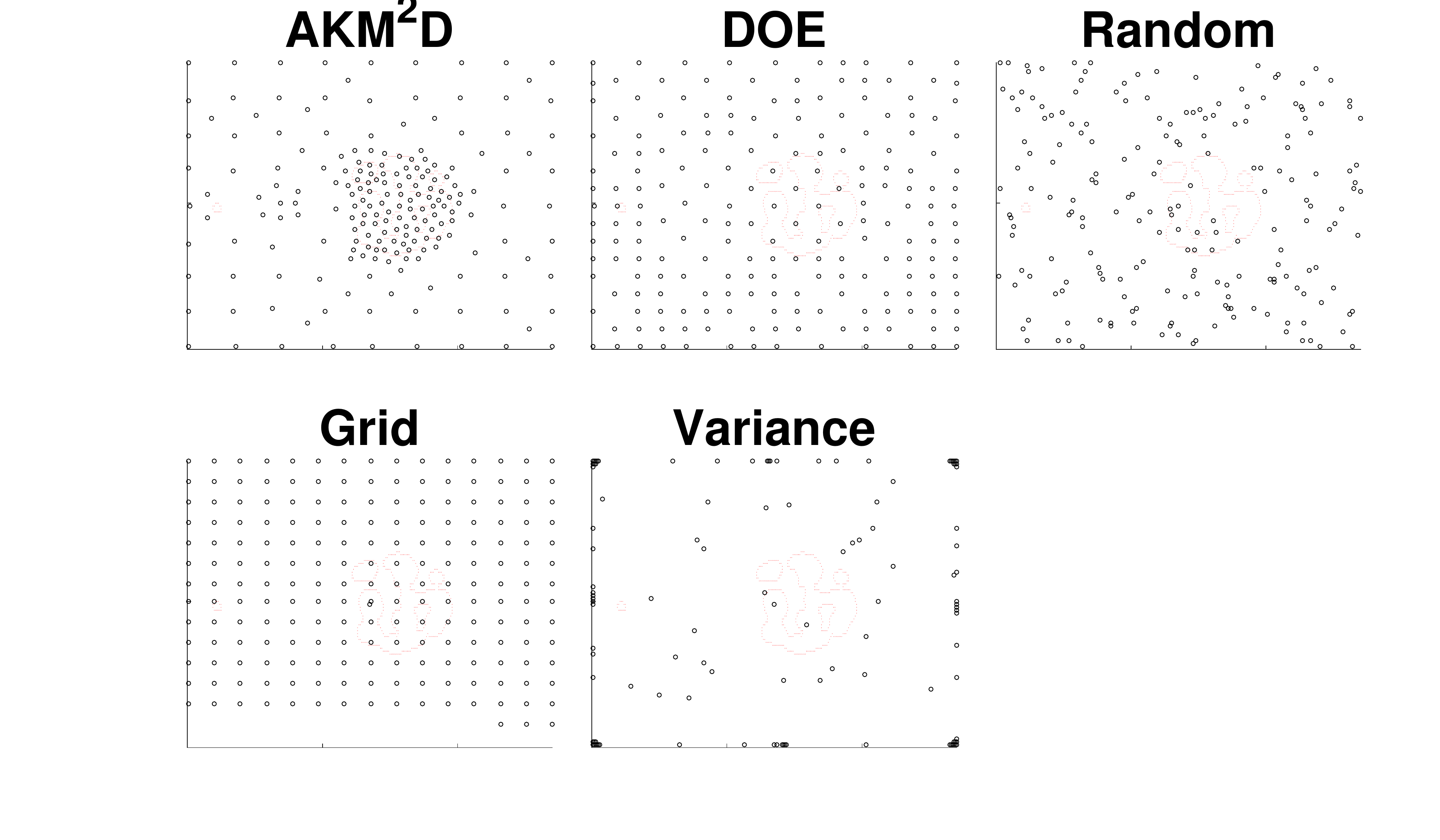}

\label{Fig: 200case}}\hfill{}\subfloat[300 sampled points]{\includegraphics[width=0.45\linewidth]{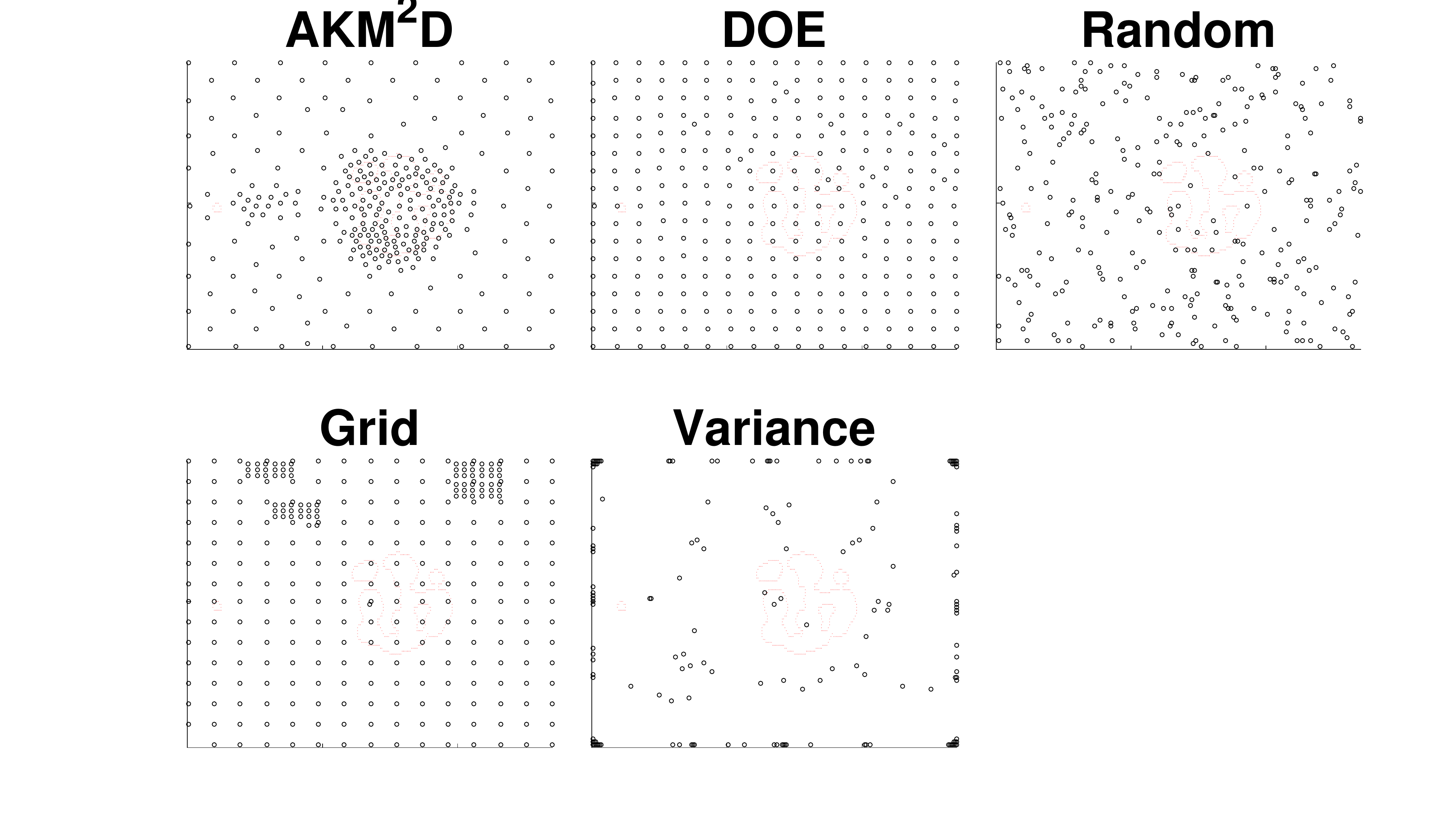}

\label{Fig: 300case}}

\caption{Sampled point pattern for all methods for 200 and 300 points}
\label{Fig: casepoint}
\end{figure}

\begin{figure}
\subfloat[200 sampled points]{\includegraphics[width=0.45\linewidth]{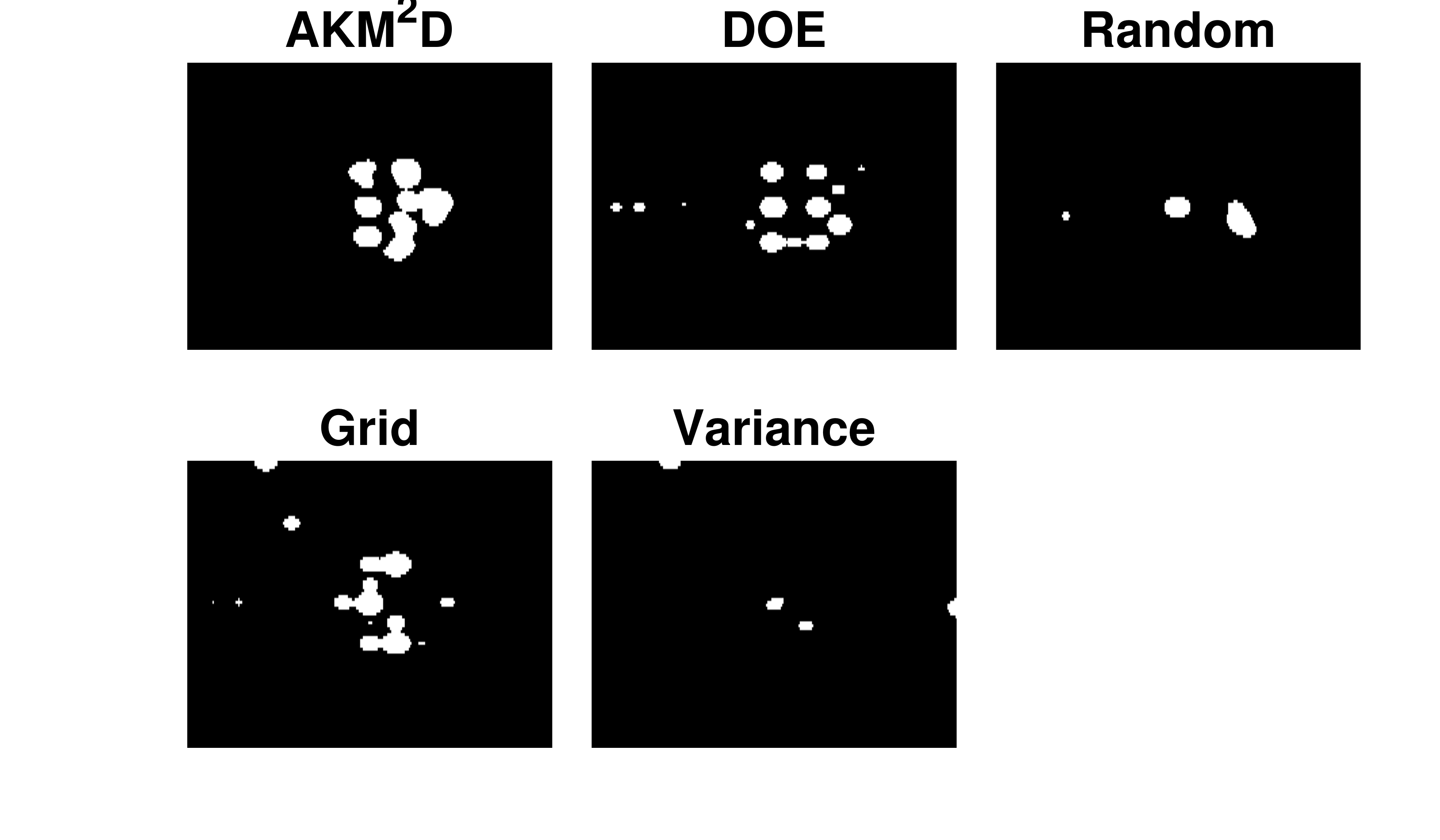}

\label{Fig: 200caseanomaly}}\hfill{}\subfloat[300 sampled points]{\includegraphics[width=0.45\linewidth]{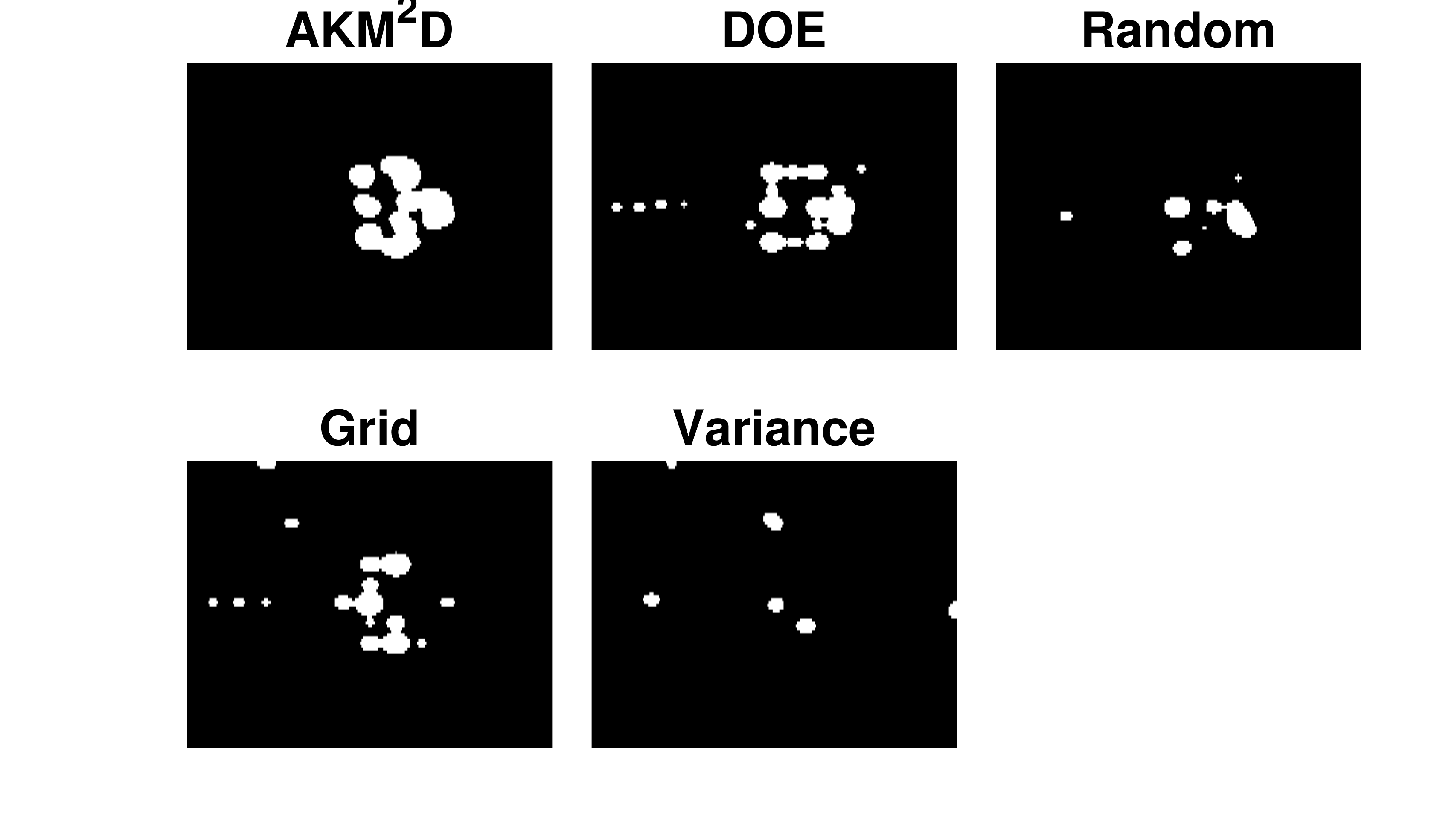}

\label{Fig: 300caseanomaly}}

\caption{Anomaly estimation result for all methods for 200 and 300 points}
\label{Fig: caseanomaly}
\end{figure}

The pattern of sampled points and detected anomalous regions by using
$200$ and $300$ points are also shown in Figures \ref{Fig: casepoint}
and \ref{Fig: caseanomaly}, respectively. From these figures, it
is clear that the irregular anomalous regions can be fully explored
by the proposed $\mathbf{\mathrm{AKM^{2}D}}$ with only $200$ sampled
points ($0.7\%$ of full sampling), which can reduce the measurement
time from $4$ hours to $2$ minutes. However, using other methods,
very few sampled points are selected in the anomalous regions.

\section{Conclusion\label{sec:Conclusion}}

Adaptive sampling for clustered anomaly detection is vital in scaling
up point-based inspection systems. In this paper, we proposed a novel
methodology for real-time adaptive sampling and anomaly detection
in large sampling spaces. In our methodology, we first developed an
adaptive sampling framework, namely the $\mathbf{\mathrm{AKM^{2}D}}$,
by optimizing a composite index. We also studied the sampling properties
and showed that the proposed method is able to balance sampling between
the exploration of the entire space and the focused sampling near
anomalies. We developed efficient and recursive algorithms to determine
the location of the next sampled point by solving the optimization
problem in real time. Then, we proposed robust kernel regression and
sparse kernel regression to update the estimates of the functional
mean and the anomalous regions after a new sample is collected. In
the simulation study, we showed that the proposed $\mathrm{\mathbf{\mathrm{AKM^{2}D}}}$
outperformed existing adaptive sampling approaches, which fail to
locate and focus on anomalous regions. Finally, the proposed method
was applied to a real case study on the anomaly detection of composite
laminates via guided wavefield test. We showed that our method can
achieve a similar detection accuracy to that of the complete sampling
by sensing only $0.7\%$ of the sampled points, and hence it can significantly
reduce the inspection time.

There are several potential research directions to be investigated.
One possible direction is to extend this method to batch sampling,
in which multiple sampled points can be selected simultaneously by
the algorithm. The other direction is to extend this into a higher
dimensional sampling space (e.g., larger than 2-D) by more efficient
optimization techniques than the grid-based methods.

\section*{Appendix A: The Proof of Proposition \ref{prop: medradius}}
\begin{proof}
The function $g_{a}(r)=(K_{h}(r,r_{a})+u)\|r-r_{a}\|^{\lambda}$ only
relies on the distance $\|r-r_{a}\|$ since the Gaussian kernel $K_{h}(r,r_{a})=\frac{1}{(\sqrt{2\pi}h)^{2}}\exp(-\frac{\|r-r_{a}\|^{2}}{2h^{2}})$
can be represented as a function of $\|r-r_{a}\|$. Let us define
$d:=\|r-r_{a}\|$. Therefore, $g_{a}(r)=\tilde{g}_{a}(d)=(p_{a}\frac{1}{2\pi h^{2}}\exp(-\frac{d^{2}}{2h^{2}})+u)d^{\lambda}$.
The local optimum can be obtained by solving $\tilde{g}_{a}'(d)=0$,
which is

\[
\tilde{g}_{a}'(d)=\frac{1}{2\pi h^{4}}\exp(-\frac{d^{2}}{2h^{2}})d^{\lambda-1}(2\pi h^{4}\lambda u\exp(\frac{d^{2}}{2h^{2}})-p_{a}d^{2}+h^{2}\lambda p_{a})=0.
\]
Consequently, $2\pi h^{2}\frac{\lambda u}{p_{a}}\exp(\frac{d^{2}}{2h^{2}})+\lambda=\frac{d^{2}}{h^{2}}$,
which is equivalent to solving
\[
-\frac{\pi h^{2}\lambda u}{p_{a}}\exp(\frac{\lambda}{2})=(-\frac{d^{2}}{2h^{2}}+\frac{\lambda}{2})\exp(-\frac{d^{2}}{2h^{2}}+\frac{\lambda}{2}).
\]
The above equation can be solved analytically by Lambert W-function
$-\frac{d^{2}}{2h^{2}}+\frac{\lambda}{2}=W(-\frac{\pi h^{2}\lambda u}{p_{a}}\exp(\frac{\lambda}{2}))$.
After some simplification, we have 
\[
d_{a}^{*}=h\sqrt{\lambda-2W(-\frac{\pi h^{2}\lambda u}{p_{a}}\exp(\frac{\lambda}{2})).}
\]
\end{proof}

\section*{Appendix B: The Proof of Proposition \ref{prop:  MEDbalance}}
\begin{proof}
We first consider the case that $g(r)$ is in the neighborhood of
$r_{a}$, defined as $\mathcal{R}_{a}=\{r|\|r-r_{a}\|\leq\|r-r_{k}\|,\forall k=1,\cdots,n\}$.
Therefore, from Proposition \ref{prop: medradius}, we know
\[
\max_{r\in\mathcal{R}_{a}}g(r)=\max_{r\in\mathcal{R}_{a}}g_{a}(r)=g_{a}(d_{a}^{*})=\frac{u(d_{a}^{*})^{\lambda+2}}{(d_{a}^{*})^{2}-\lambda h^{2}}
\]
We then consider $g(r)$ in the neighborhood of other points $r_{j}$,
which $r_{j}\neq r_{a}$. 
\begin{align*}
\max_{r\in\mathcal{R}_{j},r_{j}\neq r_{a},r_{j}\in\mathcal{M}_{n}}g(r) & =(\frac{p_{a}}{2\pi h^{2}}\exp(-\frac{\|r-r_{a}\|^{2}}{2h^{2}})+u)\|r-r_{j}\|^{\lambda}\\
 & \leq u(1+\exp(-c^{2}))d^{\lambda}
\end{align*}
The last inequality holds since $\|r-r_{a}\|_{r\in R_{j}}\geq\frac{1}{2}\|r_{j}-r_{a}\|\geq c\sqrt{2h^{2}\ln(\frac{p_{a}}{2\pi h^{2}u})}$,
which means, $\frac{p_{a}}{2\pi h^{2}}\exp(-\frac{\|r-r_{a}\|^{2}}{2h^{2}})\leq u\exp(-c^{2})$.
If 
\[
d<d_{a}^{*}(\frac{(d_{a}^{*})^{2}}{2((d_{a}^{*})^{2}-\lambda h^{2})})^{\frac{1}{\lambda}}(\frac{1}{(1+\exp(-c^{2})})^{\frac{1}{\lambda}},
\]
then, $\max_{r\in\mathcal{R}_{j},j\neq a}g(r)\leq\max_{r\in\mathcal{R}_{a}}g(r)$.
Therefore, $\mathrm{argmax}_{r}g(r)$ can be found in the neighborhood
of $r_{a}$. More specifically, according to in Proposition 1, $\mathrm{argmax}_{r}g_{a}(r)=\{r|\|r-r_{a}\|=d_{a}^{*}\}$.
\end{proof}

\section*{Appendix C: Discussion about the Remark 3.}
\begin{proof}
We first consider the MED design \citep{joseph2015sequential} $\min_{D}\{\sum_{i=1}^{n-1}\sum_{j=i+1}^{n}(\frac{q(x_{i})q(x_{j})}{d(x_{i},x_{j})})^{k}\}^{1/k}$.
When $k\rightarrow\infty$ this problem becomes $\max_{D}\min_{i,j}\frac{\|x_{i}-x_{j}\|}{q(x_{i})q(x_{j})}.$
According to the conjecture in the \citep{joseph2015sequential},
if we set $q(x)=\frac{1}{\{f(x)\}^{1/(2p)}}$, when $n\rightarrow\infty$,
the sampling algorithm will converge to $f(x)$. The papers also demonstrates
that if we solve the algorithm in a greedy algorithm, with a proper
initial design, it can achieve the same limiting behavior.

To solve $\max_{D}\min_{i,j}\frac{\|x_{i}-x_{j}\|}{q(x_{i})q(x_{j})}$
adaptively, this is equivalent to at each iteration, we like to solve
$\max_{x}f(x){}^{1/(2p)}\min_{i}f(x_{i})^{1/(2p)}\|x-x_{i}\|$ iteratively.
The limiting distribution of $\{x_{i}\}$ is actually $f(x)$. Since
when $n\rightarrow\infty$, $\|x-x_{i}\|\rightarrow0$. If $f(x)$
is a continous function, this is equivalent to solve $\max_{x}f(x){}^{1/p}\min_{i}\|x-x_{i}\|$,
which is equivalent to $\mathrm{AKM^{2}D}$ with fixed $\psi(x)$
as $\max_{x}\psi(x)\min_{i}\|x-x_{i}\|^{\lambda}$, with the limiting
distribution $f(x)=\psi(x)^{p/\lambda}$.
\end{proof}

\section*{Appendix D: Iterative Soft-thresholding for Robust Kernel Regression}

We first show the equivalency of robust kernel regression and outlier
detection in the following Lemma.
\begin{lem*}
\eqref{eq:SSD} and \eqref{eq: RKR} are equivalent in the sense that
the $\mu$ solved by both formulations are the same. 
\begin{equation}
\min_{a,\mu}\|z-a-\mu\|_{2}^{2}+\gamma\|a\|_{1}+\lambda\|\mu\|_{H}\label{eq:SSD}
\end{equation}
\end{lem*}
The detailed proof of this is shown in \citep{Mateos2012}.

It is straightforward to show that if $\mu$ is given, $a$ can be
solved by soft-thresholding as $a=S(z-\mu,\frac{\gamma}{2})$, where
$S(x,\frac{\gamma}{2})=\mathrm{sgn}(x)(\left|x\right|-\frac{\gamma}{2})_{+}$
is the soft-thresholding operator. $\mathrm{sgn}(x)$ is the sign
function and $x_{+}=\max(x,0)$. This Lemma relates the robust kernel
regression with outlier detection problem, which also explains why
$\frac{\gamma}{2}$ is a natural threshold for a point to be considered
as an outlier. Furthermore, given $a$, $\mu$ can be solved via $\mu=H(z-a)$,
where $H$ is the projection matrix computed by $H=K(K+\lambda I)^{-1}$.

\begin{algorithm} 
\caption{Optimization algorithm for Robust Kernel Regression} 
\SetKwBlock{Initialize}{initialize}{end}
\Initialize{

Choose a basis for the background as $B$

$a^{(0)}=0$

$H=K(K+\lambda I)^{-1}$

}

\While{$\ensuremath{|\mu^{(t-1)}-\mu^{(t)}|>\epsilon}$}{

Update $\mu^{(t+1)}$ via $\mu^{(t+1)}=H(z-a^{(t)})$

Update $a^{(t+1)}$ by $a^{(t+1)}=S(z-\mu^{(t+1)},\frac{\gamma}{2}))$

}

\label{alg: BCD}

\end{algorithm}

\section*{Appendix E: Additional Sensitivity Analysis}

We also perform some additional sensitivity analysis for how different
combinations of $u,\lambda$ and $h$ can affect the MMD and AMMD
for one anomalous circle region with different radius. The results
are shown in Table 2.

\begin{table}
\caption{Sensitivity Analysis for differnt anomaly sizes}

\begin{turn}{90}
\begin{tabular}{|c|c|c|c|c|c|c|c|c|c|c|c|c|c|c|}
\hline 
\multirow{2}{*}{Size} & \multirow{2}{*}{Criterion} & $h_{a}$ & \multicolumn{4}{c|}{$0.015$} & \multicolumn{4}{c|}{0.02} & \multicolumn{4}{c|}{$0.03$}\tabularnewline
\cline{3-15} \cline{4-15} \cline{5-15} \cline{6-15} \cline{7-15} \cline{8-15} \cline{9-15} \cline{10-15} \cline{11-15} \cline{12-15} \cline{13-15} \cline{14-15} \cline{15-15} 
 &  & $u$, $\lambda$ & 20 & 10 & 6.7 & 5 & 20 & 10 & 6.7 & 5 & 20 & 10 & 6.7 & 5\tabularnewline
\hline 
\hline 
\multirow{8}{*}{$0.0311$} & \multirow{4}{*}{MMD} & $10^{-13}$ & 0.0600 & 0.0600 & 0.0600 & 0.0600 & 0.0602 & 0.0600 & 0.0600 & 0.0600 & 0.0602 & 0.0602 & 0.0600 & 0.0600\tabularnewline
\cline{3-15} \cline{4-15} \cline{5-15} \cline{6-15} \cline{7-15} \cline{8-15} \cline{9-15} \cline{10-15} \cline{11-15} \cline{12-15} \cline{13-15} \cline{14-15} \cline{15-15} 
 &  & $10^{-11}$ & 0.0831 & 0.0606 & 0.0600 & 0.0600 & 0.0856 & 0.0621 & 0.0601 & 0.0600 & 0.0976 & 0.0761 & 0.0602 & 0.0600\tabularnewline
\cline{3-15} \cline{4-15} \cline{5-15} \cline{6-15} \cline{7-15} \cline{8-15} \cline{9-15} \cline{10-15} \cline{11-15} \cline{12-15} \cline{13-15} \cline{14-15} \cline{15-15} 
 &  & $10^{-9}$ & 0.1234 & 0.0881 & 0.0602 & 0.0600 & 0.1234 & 0.0958 & 0.0631 & 0.0600 & 0.1234 & 0.1151 & 0.0774 & 0.0601\tabularnewline
\cline{3-15} \cline{4-15} \cline{5-15} \cline{6-15} \cline{7-15} \cline{8-15} \cline{9-15} \cline{10-15} \cline{11-15} \cline{12-15} \cline{13-15} \cline{14-15} \cline{15-15} 
 &  & $10^{-7}$ & 0.1234 & 0.1234 & 0.0851 & 0.0600 & 0.1234 & 0.1234 & 0.0884 & 0.0602 & 0.1234 & 0.1234 & 0.1133 & 0.0602\tabularnewline
\cline{2-15} \cline{3-15} \cline{4-15} \cline{5-15} \cline{6-15} \cline{7-15} \cline{8-15} \cline{9-15} \cline{10-15} \cline{11-15} \cline{12-15} \cline{13-15} \cline{14-15} \cline{15-15} 
 & \multirow{4}{*}{AMMD} & $10^{-13}$ & 0.0216 & 0.0295 & 0.0425 & 0.0480 & 0.0210 & 0.0261 & 0.0330 & 0.0424 & 0.0206 & 0.0258 & 0.0312 & 0.0424\tabularnewline
\cline{3-15} \cline{4-15} \cline{5-15} \cline{6-15} \cline{7-15} \cline{8-15} \cline{9-15} \cline{10-15} \cline{11-15} \cline{12-15} \cline{13-15} \cline{14-15} \cline{15-15} 
 &  & $10^{-11}$ & 0.0109 & 0.0112 & 0.0219 & 0.0481 & 0.0109 & 0.0112 & 0.0202 & 0.0331 & 0.0112 & 0.0147 & 0.0184 & 0.0306\tabularnewline
\cline{3-15} \cline{4-15} \cline{5-15} \cline{6-15} \cline{7-15} \cline{8-15} \cline{9-15} \cline{10-15} \cline{11-15} \cline{12-15} \cline{13-15} \cline{14-15} \cline{15-15} 
 &  & $10^{-9}$ & 0.0071 & 0.0077 & 0.0112 & 0.0384 & 0.0077 & 0.0088 & 0.0112 & 0.0260 & 0.0107 & 0.0107 & 0.0135 & 0.0216\tabularnewline
\cline{3-15} \cline{4-15} \cline{5-15} \cline{6-15} \cline{7-15} \cline{8-15} \cline{9-15} \cline{10-15} \cline{11-15} \cline{12-15} \cline{13-15} \cline{14-15} \cline{15-15} 
 &  & $10^{-7}$ & 0.0071 & 0.0071 & 0.0071 & 0.0187 & 0.0071 & 0.0071 & 0.0071 & 0.0159 & 0.0094 & 0.0094 & 0.0094 & 0.0147\tabularnewline
\hline 
\multirow{8}{*}{$0.0517$} & \multirow{4}{*}{MMD} & $10^{-13}$ & 0.0601 & 0.0600 & 0.0600 & 0.0600 & 0.0602 & 0.0601 & 0.0600 & 0.0600 & 0.0606 & 0.0602 & 0.0600 & 0.0600\tabularnewline
\cline{3-15} \cline{4-15} \cline{5-15} \cline{6-15} \cline{7-15} \cline{8-15} \cline{9-15} \cline{10-15} \cline{11-15} \cline{12-15} \cline{13-15} \cline{14-15} \cline{15-15} 
 &  & $10^{-11}$ & 0.0861 & 0.0637 & 0.0602 & 0.0600 & 0.0882 & 0.0679 & 0.0602 & 0.0600 & 0.1110 & 0.0802 & 0.0602 & 0.0600\tabularnewline
\cline{3-15} \cline{4-15} \cline{5-15} \cline{6-15} \cline{7-15} \cline{8-15} \cline{9-15} \cline{10-15} \cline{11-15} \cline{12-15} \cline{13-15} \cline{14-15} \cline{15-15} 
 &  & $10^{-9}$ & 0.1479 & 0.0983 & 0.0607 & 0.0600 & 0.1479 & 0.1170 & 0.0691 & 0.0600 & 0.1479 & 0.1231 & 0.0823 & 0.0602\tabularnewline
\cline{3-15} \cline{4-15} \cline{5-15} \cline{6-15} \cline{7-15} \cline{8-15} \cline{9-15} \cline{10-15} \cline{11-15} \cline{12-15} \cline{13-15} \cline{14-15} \cline{15-15} 
 &  & $10^{-7}$ & 0.1479 & 0.1479 & 0.0917 & 0.0601 & 0.1479 & 0.1479 & 0.0968 & 0.0602 & 0.1479 & 0.1479 & 0.1222 & 0.0618\tabularnewline
\cline{2-15} \cline{3-15} \cline{4-15} \cline{5-15} \cline{6-15} \cline{7-15} \cline{8-15} \cline{9-15} \cline{10-15} \cline{11-15} \cline{12-15} \cline{13-15} \cline{14-15} \cline{15-15} 
 & \multirow{4}{*}{AMMD} & $10^{-13}$ & 0.0224 & 0.0308 & 0.0425 & 0.0475 & 0.0221 & 0.0277 & 0.0353 & 0.0427 & 0.0214 & 0.0261 & 0.0327 & 0.0428\tabularnewline
\cline{3-15} \cline{4-15} \cline{5-15} \cline{6-15} \cline{7-15} \cline{8-15} \cline{9-15} \cline{10-15} \cline{11-15} \cline{12-15} \cline{13-15} \cline{14-15} \cline{15-15} 
 &  & $10^{-11}$ & 0.0112 & 0.0151 & 0.0227 & 0.0475 & 0.0112 & 0.0146 & 0.0212 & 0.0355 & 0.0135 & 0.0156 & 0.0196 & 0.0310\tabularnewline
\cline{3-15} \cline{4-15} \cline{5-15} \cline{6-15} \cline{7-15} \cline{8-15} \cline{9-15} \cline{10-15} \cline{11-15} \cline{12-15} \cline{13-15} \cline{14-15} \cline{15-15} 
 &  & $10^{-9}$ & 0.0093 & 0.0107 & 0.0124 & 0.0329 & 0.0101 & 0.0109 & 0.0138 & 0.0260 & 0.0112 & 0.0112 & 0.0147 & 0.0224\tabularnewline
\cline{3-15} \cline{4-15} \cline{5-15} \cline{6-15} \cline{7-15} \cline{8-15} \cline{9-15} \cline{10-15} \cline{11-15} \cline{12-15} \cline{13-15} \cline{14-15} \cline{15-15} 
 &  & $10^{-7}$ & 0.0095 & 0.0095 & 0.0107 & 0.0224 & 0.0095 & 0.0095 & 0.0107 & 0.0175 & 0.0109 & 0.0109 & 0.0112 & 0.0156\tabularnewline
\hline 
\multirow{8}{*}{$0.0747$} & \multirow{4}{*}{MMD} & $10^{-13}$ & 0.0603 & 0.1044 & 0.1582 & 0.1582 & 0.0648 & 0.1231 & 0.1582 & 0.1582 & 0.0674 & 0.1251 & 0.1582 & 0.1582\tabularnewline
\cline{3-15} \cline{4-15} \cline{5-15} \cline{6-15} \cline{7-15} \cline{8-15} \cline{9-15} \cline{10-15} \cline{11-15} \cline{12-15} \cline{13-15} \cline{14-15} \cline{15-15} 
 &  & $10^{-11}$ & 0.0602 & 0.0826 & 0.1241 & 0.1582 & 0.0602 & 0.0859 & 0.1334 & 0.1582 & 0.0612 & 0.0957 & 0.1582 & 0.1582\tabularnewline
\cline{3-15} \cline{4-15} \cline{5-15} \cline{6-15} \cline{7-15} \cline{8-15} \cline{9-15} \cline{10-15} \cline{11-15} \cline{12-15} \cline{13-15} \cline{14-15} \cline{15-15} 
 &  & $10^{-9}$ & 0.0600 & 0.0606 & 0.0843 & 0.1165 & 0.0602 & 0.0610 & 0.0857 & 0.1233 & 0.0602 & 0.0710 & 0.0952 & 0.1499\tabularnewline
\cline{3-15} \cline{4-15} \cline{5-15} \cline{6-15} \cline{7-15} \cline{8-15} \cline{9-15} \cline{10-15} \cline{11-15} \cline{12-15} \cline{13-15} \cline{14-15} \cline{15-15} 
 &  & $10^{-7}$ & 0.0600 & 0.0600 & 0.0600 & 0.0602 & 0.0600 & 0.0600 & 0.0602 & 0.0646 & 0.0600 & 0.0601 & 0.0612 & 0.0775\tabularnewline
\cline{2-15} \cline{3-15} \cline{4-15} \cline{5-15} \cline{6-15} \cline{7-15} \cline{8-15} \cline{9-15} \cline{10-15} \cline{11-15} \cline{12-15} \cline{13-15} \cline{14-15} \cline{15-15} 
 & \multirow{4}{*}{AMMD} & $10^{-13}$ & 0.0245 & 0.0167 & 0.0150 & 0.0163 & 0.0253 & 0.0180 & 0.0176 & 0.0168 & 0.0267 & 0.0183 & 0.0179 & 0.0184\tabularnewline
\cline{3-15} \cline{4-15} \cline{5-15} \cline{6-15} \cline{7-15} \cline{8-15} \cline{9-15} \cline{10-15} \cline{11-15} \cline{12-15} \cline{13-15} \cline{14-15} \cline{15-15} 
 &  & $10^{-11}$ & 0.0310 & 0.0206 & 0.0158 & 0.0169 & 0.0299 & 0.0203 & 0.0176 & 0.0168 & 0.0290 & 0.0204 & 0.0179 & 0.0184\tabularnewline
\cline{3-15} \cline{4-15} \cline{5-15} \cline{6-15} \cline{7-15} \cline{8-15} \cline{9-15} \cline{10-15} \cline{11-15} \cline{12-15} \cline{13-15} \cline{14-15} \cline{15-15} 
 &  & $10^{-9}$ & 0.0427 & 0.0257 & 0.0221 & 0.0190 & 0.0379 & 0.0239 & 0.0180 & 0.0170 & 0.0314 & 0.0251 & 0.0199 & 0.0184\tabularnewline
\cline{3-15} \cline{4-15} \cline{5-15} \cline{6-15} \cline{7-15} \cline{8-15} \cline{9-15} \cline{10-15} \cline{11-15} \cline{12-15} \cline{13-15} \cline{14-15} \cline{15-15} 
 &  & $10^{-7}$ & 0.0502 & 0.0502 & 0.0466 & 0.0297 & 0.0433 & 0.0400 & 0.0303 & 0.0249 & 0.0425 & 0.0355 & 0.0257 & 0.0242\tabularnewline
\hline 
\end{tabular}
\end{turn}

\label{table: Sensitivity}
\end{table}

From these tables we can conclude that a larger $u$ will lead to
a better exploration of the entire background (a smaller MMD) but
lead to worse exploitation of the anomaly (a larger AMMD). If $u=10^{-7}$,
we find that the MMD always stays at $0.06$. The reason is that the
algorithm actually trapped in exploitation since it violates the inequality
$(4\pi u\exp(\frac{\lambda}{2}))^{-1/\lambda}h\sqrt{\lambda}<1$,
provided in the Proposition 2.

\section*{Appendix F: Sampling Behavior in a Video}

\begin{figure}
\begin{centering}
\includegraphics[width=0.5\linewidth]{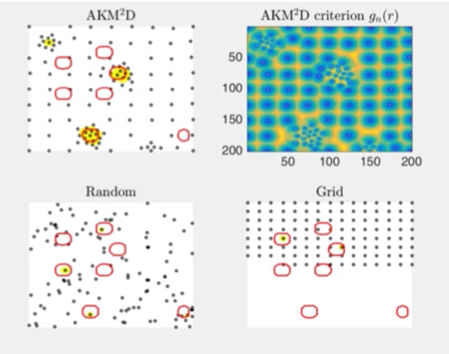}
\par\end{centering}
\label{Fig: snapshot}

\caption{A snapshot of the video showing the sampling behavior
balance between exploration and exploitation}
\end{figure}

This appendix is to illustrate how the algorithm behaves and balance
between exploration and exploitation via a video in the supplementary
file. A snapshot of the video is shown in Figure \eqref{Fig: snapshot}.
The first video (i.e., upper left) shows sensing points (i.e., in
black dots), true anomaly (i.e., the boundary is shown in the red
curve), and estimated anomalies (i.e., estimated anomalies). The second
video (i.e., upper right) shows the sampling criterion function, where
the next point tends to select the point with larger value (i.e.,
the yellow regions). The third and fourth videos show the performance
of the benchmark methods. It can be clearly seen that the algorithm
will alternatively sample near the red curve (i.e., exploitation)
and in the background (i.e., exploration).

\bibliographystyle{plainnat}
\bibliography{adaptive}

\end{document}